\documentclass[runningheads]{llncs}
\usepackage[T1]{fontenc}
\usepackage{graphicx}
\usepackage{algorithm}
\usepackage[noend]{algpseudocode}
 \usepackage{amsmath}
  \usepackage{amssymb}

\newcommand{\half}{\frac{1}{2}}

\newtheorem{thm}{Theorem}

\renewcommand{\P}{\mathbb{P}}
\newcommand{\E}{\mathbb{E}}

\begin{document}
\title{sqSGD: Locally Private and Communication Efficient Federated Learning}
%
%
\author{Yan Feng\inst{1,5} \and
Tao Xiong\inst{2} \and
Ruofan Wu\inst{3} \and
Lingjuan Lv \inst{4} \and
Leilei Shi \inst{6}}
\authorrunning{F. Author et al.}
\institute{Tsinghua Univerisity \and Google \and Ant Group \and
Sony AI \and Meituan \and Galixir
}
\maketitle              
\begin{abstract}
Federated learning (FL) is a technique that trains machine learning models from decentralized data sources. We study FL under local notions of privacy constraints, which provides strong protection against sensitive data disclosures via obfuscating the data before leaving the client. We identify two major concerns in designing practical privacy-preserving FL algorithms: communication efficiency and high-dimensional compatibility. We then develop a gradient-based learning algorithm called \emph{sqSGD} (selective quantized stochastic gradient descent) that addresses both concerns. The proposed algorithm is based on a novel privacy-preserving quantization scheme that uses a constant number of bits per dimension per client. Then we improve the base algorithm in three ways: first, we apply a gradient subsampling strategy that simultaneously offers better training performance and smaller communication costs under a fixed privacy budget. Secondly, we utilize randomized rotation as a preprocessing step to reduce quantization error. Thirdly, an adaptive gradient norm upper bound shrinkage strategy is adopted to improve accuracy and stabilize training. Finally, the practicality of the proposed framework is demonstrated on benchmark datasets. Experiment results show that sqSGD successfully learns large models like LeNet and ResNet with local privacy constraints. In addition, with fixed privacy and communication level, the performance of sqSGD significantly dominates that of various baseline algorithms.

\keywords{Differential privacy  \and Communication efficiency \and Federated learning.}
\end{abstract}
\section{Introduction}
Federated learning (FL) \cite{kairouz2019advances,konevcny2016federated} is a rapidly evolving application of distributed optimization to large-scale learning or estimation scenarios where multiple entities. called \emph{clients}, collaborate in solving a machine learning problem, under the coordination of a \emph{central server}. Each client’s raw data is stored locally and not exchanged or transferred. To achieve the learning objective, the server collects minimal information from the clients for immediate aggregation. FL is particularly suitable for mobile and edge device applications since the (sensitive) individual data never directly leave the device and has seen deployments in industries  ~\cite{FL,hard2019federated,leroy2019fed}. While FL offers significant practical privacy improvements over centralizing all the training data, it lacks a formal privacy guarantee. As discussed in \cite{melis2018inference}, even if only model updates (i.e. gradient updates) are transmitted, it is easy to compromise the privacy of individual clients. \par

Differential privacy (DP) \cite{dwork2014algorithmic} is the state-of-the-art approach to address information disclosure. Differentially private algorithms fuse participation of any individual via injecting algorithm-specific random noise. In FL settings, DP is suitable for protecting against \emph{external adversaries}, i.e. a malicious analyst that tries to infer individual data via observing final or intermediate model results. However, DP paradigms typically assume a \emph{trusted curator}~\cite{dwork2014algorithmic}, which corresponds to the server in the FL setting~\cite{brendan2018learning,lyu2020privacy}. This assumption is often not satisfied in practical cases, under which users that act as clients may not trust the service provider that acts as the server. Therefore, local notions of privacy that provides privacy protection on the \emph{individual level} are often preferred in FL scenarios \cite{kairouz2019advances,bhowmick2018protection}. Local differential privacy (LDP) \cite{kasiviswanathan2011can,dwork2014algorithmic} is the most natural choice of local privacy model. Algorithms that satisfy LDP constraints typically operate via applying randomized mechanisms that obfuscate the data before leaving the client\cite{dwork2014algorithmic}. In terms of FL, the most representative operation is to optimize the learning objective via stochastic gradient (SGD) methods~\cite{mcmahan2017communication}. Therefore, most of the existent private FL frameworks have focused on the design of private SGD algorithms, which could be treated as a special case of distributed mean estimation under privacy constraints \cite{agarwal2018cpsgd,g2019vqsgd,chen2020breaking}. \par 

Despite its decent protection guarantee under reasonably low privacy budgets, adhering to the LDP protocol makes the learning problem much harder compared to the non-private case. Previous works \cite{duchi2018minimax,duchi2018right} have established fundamental results that heuristically state "local differential privacy makes easy problems not that easy". The most relevant consequence with respect to FL is best understood using the minimax lower bounds for private mean estimation \cite{duchi2018minimax}: the result suggests that in order to satisfy the LDP constraint, one must sacrifice a multiplicative accuracy penalty that is linear in $d$ -- the dimension of the estimation problem. For modern deep learning scenarios that typically involve a model with a very high dimensional parameter (i.e., to the order of millions), this would imply too high an estimation cost to model learning.

Yet another important challenge of FL is communication efficiency. For each round, each client transmit $ O(d\delta) $ bits to the server, where $ \delta $ is the minimum number of bits to represent a real number with desired precision. The resulting communication cost
\footnote{
	In general, there are two sources of communication cost in a distributed learning scenario that requires sequential interactions between the server and the clients: downlink cost that happens during clients downloading the updated model from the central server, and uplink cost that happens when clients sending their updates to the server. In was previously noted in \cite{konevcny2016federated} that uplink cost usually dominates downlink costs in FL settings. Hence in the scope of this work, we will refer to communication cost as uplink communication cost. 
}
is usually not affordable for mobile scenarios. In fact, the primary bottleneck for FL is usually communication, especially for deep neural architectures \cite{kairouz2019advances}. It is thus of significant importance to reduce the communication cost to a reasonable level. Overall, we identify two main challenges in building practical privacy-preserving FL algorithms: (1) Communication Efficiency: The algorithm shall be communication efficient, measured in terms of bits transferred per client in a single round; (2) High-dimensional compatibility: The algorithm shall be able to handle models with high-dimensional inputs with tolerable performance degradation. 

\textbf{Summary of contributions.} In this paper, we propose a practical solution to locally private FL setting in order to address the above concerns, which we term \emph{selective quantized stochastic gradient descent (sqSGD)}. Our contributions are summarized as follows:
\begin{enumerate}
	\item We propose a novel algorithm for locally private multivariate mean estimation. The proposed algorithm, $ \text{\textsf{PrivQuant}}_{K, \infty} $, stochastically quantizes each dimension of each sample to $ K $ equally spaced levels, and is shown to satisfy both LDP and reconstruction protection guarantee \cite{bhowmick2018protection}. As a base algorithm, $ \text{\textsf{PrivQuant}}_{K, \infty} $ offers controllable communication cost and provides decent privacy protection against reconstruction  even when a high privacy budget is used. 
	\item We make several improvements to the basic algorithm under the gradient-based learning scenario. Specifically, we attribute the estimation error of the base algorithm to perturbation error caused by privacy constraints, and quantization error. Then we utilize a gradient subsampling strategy to simultaneously reduce communication cost and improve private estimation utility, and a randomized rotation operation to reduce quantization error. Moreover, we further improve the training performance via adopting an adaptive gradient norm upper bound shrinkage strategy.
	\item We verify the performance of sqSGD on benchmark datasets using standard neural architectures like LeNet \cite{LeNet} and ResNet \cite{he2016deep}, and make systematic studies on the impact of both communication and privacy constraints. Specifically, under the same privacy and communication constraints, our model outperforms several baseline algorithms by a significant margin. 
\end{enumerate}

\section{Preliminary}\label{sec: meth}
\subsection{Federated learning}
We will be focusing on distributed learning procedures that aim to solve an empirical risk minimization problem in a decentralized fashion:
\begin{align}\label{erm}
	\min_{\theta \in \Theta} L(\theta), \quad L(\theta) =  \frac{1}{N}  \sum_{m = 1}^{M}L_m(\theta)
\end{align}
where $ M $ denotes the number of clients. Let $ Y_i $ denotes training data (inputs as well as the labels) that belongs to the $ m $-th client consisting $ N_m $ data points, and $ N = \sum_{m=1}^M N_m $. The term $ L_m(\theta) = \sum_{y \in Y_m} \ell(\theta, y) $ stands for locally aggregated loss evaluated at parameter $ \theta $, where $ \ell: \Theta \times \mathcal{Y} \mapsto \mathbb{R}_+ $ is the loss function depending on the problem context and we use $ \Theta $ and $ \mathcal{Y} $ to describe parameter and data domain respectively. We will seek to optimize \eqref{erm} via gradient-based methods. In what follows we will use $ [n], n \in \mathbb{N}_+ $ to denote the set $ \{1, 2, \ldots, n\} $. At round $ t $, the procedure iterates as:
\begin{enumerate}
	\item Server distributes the current value $ \theta_t $ among a subset $ S $ of clients. 
	\item For each client $ s \in S $, a local update is computed $ \Delta_s = g(\theta_t, Y_s) $ and transmitted to the server.
	\item The server aggregates all $ \{\Delta_s\}_{s \in S} $ to obtain a global update $ \Delta $ and updates the parameter as $ \theta_{t+1} = \theta_t + \Delta $.
\end{enumerate}
We distinguish two practical scenarios as \emph{cross-silo} FL and \emph{cross-device} FL \cite{kairouz2019advances}. In cross-silo FL, $ M $ is relatively small (i.e. $ M \le 100 $ ) and usually each client has a moderate or large amount of data (i.e. $ \min_{m} N_m \gg 1 $). As a consequence, all the clients participate in the learning process (i.e. $ S = [M] $). In each iteration a client locally computes a negative stochastic gradient $ g(\theta_t, Y_s) = -\frac{1}{R}\sum_{y \in \mathcal{R}_s} \nabla \ell(\theta_t, y) $ of $ L_s(\theta_t)/N_s $, based on a uniform random subsample $ \mathcal{R}_s \subset  Y_s $ of size $ R $. In cross-device FL, $ S $ is a uniformly random subset of $ [M] $, and $ g(\theta_t, Y_s) = -\frac{1}{N_s} \sum_{y \in Y_s} \nabla \ell(\theta_t, y)$ is the average negative gradient over $ Y_s $ evaluated at $ \theta_t $. In this paper we will be focusing on the {\fontfamily{qcr}\selectfont FedSgd} aggregation protocol~\cite{mcmahan2017communication} corresponding to a stochastic gradient step with learning rate $ \eta $, i.e., $ \Delta = \eta \sum_{s \in S} \frac{N_s}{N}g(\theta_t, Y_s)$.

\subsection{Local differential privacy} 
Generally speaking, in FL, there are two sources of data that need privacy protection: the parameter $ \theta_t $ and the updates $ \{\Delta_s\}_{s \in S} $. In this paper we will be focusing on the protection of $ \{\Delta_s\}_{s \in S} $ . Note that the protection of $ \theta_t $s could be done via applying standard central DP techniques as in \cite{bhowmick2018protection}. To begin our discussion on suitably defined privacy models, we first review the local differential privacy model:
\paragraph{	Local differential privacy \cite{kasiviswanathan2011can} }we say a randomized algorithm $ \mathcal{A} $ that maps the private data $ X \in \mathcal{X} $ to some value $ Z = \mathcal{A}(X) \in \mathcal{Z} $ is \emph{$ \epsilon $-locally differentially private}, if the induced conditional probability measure $ \mathbb{P}\left( \cdot | X = x\right)  $ satisfies that for any $ x, x^\prime \in \mathcal{X} $ and any $ Z \in \mathcal{Z} $:
\begin{align}
	e^{-\epsilon} \le  \dfrac{\mathbb{P}\left( Z | X = x\right) }{\mathbb{P}\left( Z | X = x^\prime\right)} \le e^{\epsilon}
\end{align}
Under the LDP paradigm, an adversary\footnote{Throughout the paper we assume the adversaries to be \emph{semi-honest} \cite{lyu2020threats}, i.e. they are curious about individual data but follow the FL protocol correctly.}
is allowed to observe the final learned model, as well as the information about potential participants except for the precise membership list. An $ \epsilon- $ LDP algorithm would guarantee that even the adversary knows the underlying data is one of $ x $ and $ x^\prime $, he/she could not distinguish them in a probabilistic sense, i.e., the sum of type I and type II error is at least $ 1/(1+e^\epsilon) $. 

\subsection{Mean estimation under LDP: hardness results} 
LDP is a very strong privacy protection model that protects individual data against \emph{inference attacks} that aims at inferring the membership of arbitrary data from the data universe. For this level of protection to work properly, a small privacy budget is required, i.e, $ \epsilon = O(1) $. However, using a low $\epsilon$ has fundamental impacts to the estimation performance of locally private estimators \cite{duchi2019lower}. Information theorectically, privatizing an input $ X $ with privacy level $ \epsilon $ is equivalent to limiting the bits of information that is possible to communicate about $ X $ to $ \epsilon $ bits, regardless of the informativeness of $ X $. As a consequence, there are dimension dependent penalties for many locally private estimation. The one that is of particular interest in this paper is the \emph{private mean estimation} problem. In \cite{duchi2018minimax}, the authors showed that upon estimating a mean vector inside a $ d- $dimensional unit ball $ \theta \in \mathbb{B}(0, 1) $ based on $ \epsilon- $ locally differentially private views (generated via mechanism $ \mathcal{A} $) of i.i.d. random vectors with mean $ \theta $, the minimax risk under $ \ell_2 $ loss, with sample size $ N $ satisfies the lower bound:
\begin{align}\label{minimax}
	r_{\text{\textsf{me}}}(\epsilon, \ell_2) := \min_{\hat{\theta}}\max_{\theta, \mathcal{A}} \mathbb{E}\left[ \left( \hat{\theta} - \theta\right)^2   \right] \gtrsim \dfrac{d}{\epsilon^2 N} \vee 1
\end{align}
Note that the federated gradient estimation subproblems in most FL procedures are special cases of private mean estimation. The fundamental risk reflects the hardness of adopting small $ \epsilon $ private learning procedures in FL: for modern machine learning problems where the parameter dimension $ d $ is large and the sample size is typically of order $ N = O(d) $ or even $ N = o(d) $, the gradient estimation error could be of constant order, prohibiting most large models from learning successfully. 

\subsection{Protection against reconstruction} 
The fundamental limits of locally private estimation suggest that low $\epsilon$ learning is at the risk of low accuracy for high dimensional learning scenarios. To mitigate this issue, \cite{bhowmick2018protection} proposed a reasonably weakened privacy model that is achievable using high $ \epsilon $ LDP algorithms. The intuition is that allowing adversaries with the capability of carrying membership inference attack may be overly pessimistic: to conduct an effective inference attack the adversary shall come up with "the true data" and only use the privatized output to verify his/her belief about the membership. If the prior information is reasonably constrained for the adversary, we may adopt a weaker, but still elegant type of privacy protection paradigm that protects against \emph{reconstruction attacks}. Heuristically, reconstruction attack aims at recovering individual data with respect to a well-defined criterion, given a prior over the data domain that is not too concentrated. Formally, we adopt the definition in \cite{bhowmick2018protection}:

\textbf{Reconstruction attack \cite{bhowmick2018protection}} Let $ \pi $ be a prior distribution over the data domain $ \mathcal{Y} $ that encodes the adversary's prior belief. We describe the generation process of privatized user data as
\begin{align*}
	Y \rightarrow X \rightarrow Z=\mathcal{A}(X)
\end{align*}
for some (privacy-protecting) mechanism $ \mathcal{A} $. Additionally let $ f: \mathcal{Y} \mapsto \mathbb{R}^k $ be the target of reconstruction (i.e., the adversary wants to evaluate the value $ f(Y) $) and $ L_{\text{\textsf{rec}}}: \mathbb{R}^k \times \mathbb{R}^k \mapsto \mathbb{R}_+ $ be the reconstruction loss serving as the criterion. Then an estimator $ \zeta: \mathcal{X} \mapsto \mathbb{R}^k $ provides an $ (\alpha, p, f) $-reconstruction breach for the loss $ L_{\text{\textsf{rec}}} $ if there \emph{exists} some $ z \in \mathcal{Z} $ such that:
\begin{align}
	\mathbb{P}\left( L_{\text{\textsf{rec}}}(f(X), \zeta(z)) \le \alpha | \mathcal{A}(X) = z\right) > p
\end{align}
The existence of a reconstruction breach provides an attack that effectively breaks the privatized algorithm for some output $ z $. Hence to provide protection against reconstruction attacks, we need the following to hold:
\begin{align}\label{eqn: protection_reconstruction}
	\sup_\zeta \sup_{z \in \mathcal{Z}} 	\mathbb{P}\left( L_{\text{\textsf{rec}}}(f(X), \zeta(z)) \le \alpha | \mathcal{A}(X) = z\right) \le p
\end{align}
When \eqref{eqn: protection_reconstruction} holds, the mechanism $ \mathcal{A} $ is said to be $ (\alpha, p, f) $-protected against reconstruction for the loss $ L_{\text{\textsf{rec}}} $. In \cite[Lemma 2.2]{bhowmick2018protection}, the authors 
pointed out that local privacy as traditionally employed may prove too stringent for practical use, especially in modern high-dimensional statistical and machine learning problems. Their obtained results suggested that using LDP mechanisms with \emph{large} $ \epsilon $ may still provide decent protection against reconstruction. Hence for the rest of the paper, we focus on \emph{developing LDP mechanisms with large $ \epsilon $s}, i.e., the scale of $ \epsilon $ will be chosen so that the protection against reconstruction is enough for the underlying problem. \par 

\section{Our model}
\subsection{Threat model}
For threat model of reconstruction attack, we 
follow~\cite{bhowmick2018protection}: instead of considering an adversary with access to all data, we consider "curious" onlookers, who wish to decode individuals' data but have little prior information on them. 
While this brings us away from the standard guarantees of differential privacy, we can still provide privacy guarantees for the type of onlookers we consider~\cite{bhowmick2018protection}.

\subsection{Base mechanism}
As a starting point, we base our communication efficient mean estimation algorithm on the \emph{stochastic $ k$-level quantization}\cite{suresh2017distributed} scheme: We assume a uniform upper bound $ U $ on the $ \ell_\infty $ norm of any individual gradients over the whole course of the FL process. The quantization procedure is described as follows: Let $ K \ge 2 $ be the desired quantization level. We quantize the gradient to a sequence of points $ -U = B_1 < B_2 < \ldots, < B_{K-1} < B_{K} = U $, where
\begin{align}
	B_k = -U + \dfrac{2(k - 1)U}{K - 1}
\end{align}
The case for $ K = 2 $ corresponds to using only the endpoints $ \{-U, U\} $ for quantization. We will denote the quantization range as $ \mathcal{B} = \{B_k\}_{k=1}^K $. For each dimension $ j \in [d] $ of the individual gradient $ X^j $, we first locate $ X^j $ to one of the $ K $ bins via finding a $ k^* $ such that $ X^j \in [B_{k^*}, B_{k^* + 1}) $ and round $ X^j $ to the boundary of the bin as:
\begin{align}\label{quantization}
	\widehat{X}^j = 
	\left\{
	\begin{array}{@{}l@{\thinspace}l}
		B_{k^* + 1}\quad \text{with probability } \frac{k^*(X^j - B_{k^*})}{2U} \\
		B_{k^*} \quad \text{otherwise}
	\end{array}
	\right.
\end{align}
Note that $ \widehat{X}^j $ is an unbiased estimator of $ X^j $. The quantization scheme reduces the communication cost to $ d\log_2 K $ bits per round per client. \par 

Next we privatize $ \widehat{X}^j $ by constructing a locally private unbiased estimator of $ \widehat{X}^j $ (where we fix the randomness of the quantization step), thereby obtaining a locally private estimator of $ X $ with its value resides in $ \mathcal{B}^d $. To describe the privatization scheme we introduce some new notations: for $ \forall v_1, v_2  \in \mathcal{B}^d $, define $ \mathcal{M}(v_1, v_2) = \#\{j: v_1^j = v_2^j \} - \#\{j: v_1^j \ne v_2^j \} $ where $ \#C $ stands for the number of elements in the set $ C $. For a given positive integer $ \kappa \in \{0, \ldots, d - 1\} $, let $ \overline{\mathcal{S}}(v; \mathcal{B}^d) := \{u: \mathcal{M}(u, v) > \kappa \} $ and $ \underline{\mathcal{S}}(v; \mathcal{B}^d) := \{u: \mathcal{M}(u, v) \le \kappa \} $. The privatization procedure, which we termed $\text{\textsf{PrivQuant}}_{K, \infty}$, is summarized in Algorithm\ref{alg: privQ}. To state the privacy guarantee of $\text{\textsf{PrivQuant}}_{K, \infty}$, we identify 
the setup for reconstruction attack, 
which requires additional specifications for the priors and evaluation function held by the adversary, which we state as follows: we assume the evaluation function $ f $ to be a map from $ \mathcal{Y} $ to a compact set $ \mathcal{C} \subset \mathbb{R}^r $. For any absolutely continuous prior $ \pi $ on $ \mathcal{Y} $, we use $ \pi_f $ to denote the induced prior on $ f $. Let $ \pi_0 $ be the uniform prior over $ \mathcal{C} $, we will be focusing on priors that belong to the following set:
\begin{align}
	\mathcal{P}_f(\rho_0) = \left\lbrace \pi: \sup_{y \in \mathcal{C}}\log\dfrac{d \pi_f(y)}{d \pi_0 (y)} \le \rho_0 \right\rbrace 
\end{align}
where $ \rho_0 $ is a non-negative number. The set $ \mathcal{P}_f(\rho_0) $ characterizes prior beliefs that are "not much more certain than uniformly random guessing" over all possible outcomes that belong to the image of the evaluation function. We state the theorem under the \emph{orthogonal reconstruction attack} that uses the evaluation function $ f_A(x) = Ax/\left\| x\right\|_2  $ with $ A \in \mathbb{R}^{r \times d} $ an orthonormal matrix, i.e., $ AA^T = I_r $, and the reconstruction criterion is chosen as the $ \ell_2 $ error $ L_{\text{\textsf{rec}}}(x, x^\prime) = \left\| x/\left\| x\right\|_2  - x^\prime/\left\| x^\prime \right\|_2 \right\|_2^2  $. 

\begin{thm}	\label{theorem:privacy-infinity}
	For any $ K \ge 2 $, consider $\text{\textsf{PrivQuant}}_{K, \infty}$ as a mechanism parameterized by $ (\kappa, p) $. Let $\tau := {\lceil
		\frac{d+\kappa+1}{2} \rceil }$, and let $ (\kappa, p) $ be chosen such that the following relation holds:
	\begin{equation}
		\frac{p}{1 - p}
		\times
		\frac{\sum_{\ell =0}^{\tau-1 }{{d}\choose{\ell}} (K - 1)^{(d - \ell)}}{
			\sum_{\ell = \tau }^d {{d}\choose{\ell}} (K - 1)^{(d - \ell)}} \le e^\epsilon
		\label{eq:sufficient_kappaGEN}
	\end{equation}
	then $ Z $ is an unbiased estimator of $ X $, i.e. $ \mathbb{E}(Z) = X $, with randomness jointly over the quantization step and sampling step. Moreover, we have the following privacy guarantees:
	\begin{description}
		\item[Inference attack protection] $\text{\textsf{PrivQuant}}_{K, \infty}( \cdot, \kappa, p)$ is
		$\epsilon$-locally differentially private.
		\item[Reconstruction attack protection] with $ k \ge 4 $ and $ a \in [0, 1] $, $\text{\textsf{PrivQuant}}_{K, \infty}( \cdot, \kappa, p)$ is $ (\sqrt{2 - 2a}, \omega(a), f_A) $-protected against reconstruction for 
		\begin{align}\label{eqn: recontructprotection}
			\omega(a) = \sqrt{8} \exp\left( - \dfrac{(r-1)a^2}{2} \right) \exp(\epsilon + \rho_0)
		\end{align}
	\end{description}
	
\end{thm}
\begin{algorithm}
	\caption{Private $ K$-quantized $ \ell_\infty $ Ball $ \text{\textsf{PrivQuant}}_{K, \infty} $}
	\label{alg: privQ}
	\begin{algorithmic}[1]
		\Require $X \in [-U,U]^d$, $\kappa \in \{0,\cdots, d-1\}, p \ge \half$, $\tau = {\lceil \frac{d+\kappa+1}{2} \rceil }$.
		\State Quantize each coordinate of $ X $ to $ \mathcal{B} $ using \eqref{quantization} to obtain $ \widehat{X} \in \mathcal{B}^d $.
		\State Sample a random vector $ V $ as follows: with probability $ p $, $ V $ is sampled uniformly at random from $ \overline{\mathcal{S}}(\widehat{X}; \mathcal{B}^d) $; otherwise with probability $ 1 - p $, $ V $ is sampled uniformly at random from $  \underline{\mathcal{S}}(\widehat{X}; \mathcal{B}^d) $.
		\State Calculate normalizing factor
		\begin{equation*}
			m = p\frac{{d-1 \choose{\tau - 1} }(K - 1)^{d - \tau} }{\sum_{\ell = \tau}^d {{d}\choose{\ell}} (K - 1)^{d - l}}
			- (1 - p) \frac{ {{d-1}\choose{\tau -1 }} (K - 1)^{d - \tau} }{\sum_{\ell = 0}^{\tau -1 } {{d}\choose{\ell}} (K - 1)^{d-l}}
		\end{equation*}
		\Return $Z = \frac{1}{m} \cdot V$
	\end{algorithmic}
\end{algorithm}
The proof will be deferred to appendix \ref{appendix: A1}. 
Theorem \ref{theorem:privacy-infinity} implies that, if the adversary aims at reconstructing a non-vanishing fraction of the private data with $ r = O(d) $, for small $ a $ that allows coarse reconstruction, we only need $ \epsilon $ and $ \rho_0 $ to be of smaller order than $ d $ to ensure that any reconstruction attack will not succeed with reasonable probability. 
Note for $ K=2 $ levels and $ U = 1 $, Algorithm \ref{alg: privQ} reduces to algorithm $ \textsf{PrivUnit}_\infty $ in \cite{bhowmick2018protection}. The improvement of taking $ K > 2 $ could be intuitively justified as follows: the $ \ell_2 $ estimation error of $ \text{\textsf{PrivQuant}}_{K, \infty} $ depends on the $ \ell_\infty $ norm of the output vector $ Z $, which is further governed by the normalizing factor $ m $ in \ref{alg: privQ}. A larger $ m $ reduces the scale of the output vector $ Z $, thereby reducing the variance and improving efficiency. To this end, note that $ m $ is monotonically increasing in $ K $, hence using a larger quantization range improves estimation performance. 
\par 
We may view the estimation error of $ Z $ (measured in terms of $ \ell_2 $ risk) as coming from two parties: one from the quantization step, and the other from the privatization step. In the following sections we will take closer looks at the two sources of error, and develop corresponding improvements. Indeed as verified by our empirical study (see Section \ref{experiments}), certain improvements are \emph{necessary} to achieve high-dimensional compatibility.

\subsection{Improving privatization utility via selective gradient update}
$ \text{\textsf{PrivQuant}}_{K, \infty} $ provides an unbiased estimator of individual gradients, but the considerable amount of noise injected would hurt the training process. This defect is amplified in deep learning scenarios: empirically, \cite{lin2018deep} observed that for many representative deep architectures, most of the gradient dimensions are nearly sparse. Hence privatizing the whole gradient vector would result in a high privatization error that is even orders of magnitude higher than the norm of the gradient itself. \par 
The (almost) sparse structure of the gradients suggests a modification to the estimation scheme via privatizing and transmitting only a fraction of the gradient dimensions, which could be regarded as significantly reducing variance at the cost of a small amount of bias. Such techniques are closely related to \emph{gradient compression} that sends a few most significant gradient dimensions \cite{lin2018deep}. Typically, gradient compression techniques require selecting top-$ k $ gradients measured in absolute magnitude \cite{lin2018deep}. \par

\vspace{-1.5em}
\begin{algorithm}
    \footnotesize
	\caption{sqSGD: Selective Quantized SGD with local privacy guarantee}
	\label{alg: privSQ}
	\begin{algorithmic}[1]
		\Require Training data $ \{Y_i\}_{i=1}^M $, initial gradient norm bound $ U$, privacy budget $ \epsilon = \epsilon_1 + \epsilon_2 $, sampling ratio $ r \in (0, 1) $, local correction factor $ \alpha, \beta$, learning rate $ \eta $, batch size $ B $.
		\State Find a $ (\kappa_{\epsilon_1}, p_{\epsilon_1}) $ pair that satisfies relation \eqref{eq:sufficient_kappaGEN}
		\State Calculate the corresponding normalizing constant $ m_\epsilon $ as in Algorithm \ref{alg: privQ}
		\State Set $ \tilde{d} = 2^{\lfloor \log_2(rd)\rfloor} $, and using public randomness to generate a random matrix $ R \in \mathbb{R}^{\tilde{d} \times \tilde{d}} $, according to the construction in \eqref{hadamard}
		\State Initialize model parameter $ \theta_0 $ and local residuals $ \text{\sffamily res}_{i, 0} = \mathbf{0}, \forall i = 1, \ldots, N $
		\For{$ t = 0, \ldots, T-1 $}
		\State Sample a batch of indices $ S_t \in [N] $ with size $ B $
		\State Server send $ (U_t, \theta_t) $ to all the selected clients
		\For{$ s \in S $}
		\State {\fontfamily{qcr}\selectfont //* Client Side Update *//}
		\State Select a uniformly random subset of $ [d] $ with
		\State cardinality $ \tilde{d} $, denoted as $ \mathcal{D}_s $, also let $ \mathcal{D}_s^C $ denote $ [d]\backslash \mathcal{D} $.
		\State Calculate local (stochastic) gradient $ X_{s, t} = g(Y_{s, t}, \theta_t) $
		\State Clip the local gradient to the $ \ell_2 $ ball with radius $ U $
		\State Accumulate and Apply rotation 
		\State $ \tilde{X}_{s, t}= R\left( \text{\sffamily res}_{i, t}[\mathcal{D}_s] + \beta X_{s, t}[\mathcal{D}_s] \right)  $
		\State Rotate and update local residual 
		\State $ \text{\sffamily res}_{s, t+1}[\mathcal{D}_s^C] =  \text{\sffamily res}_{s, t}[\mathcal{D}_s^C] + \alpha X_{s, t}[\mathcal{D}_s^C] $ \State and $ \text{\sffamily res}_{s, t+1}[\mathcal{D}_s] = 0 $
		\State Calculate $Z_{s, t} = \text{\textsf{PrivQuant}}_{K, \infty}(\tilde{X}_{s, t}, \kappa_{\epsilon_1}, p_{\epsilon_1})$
		\State Calculate $ U_{s, t} = \text{\textsf{ScalarDP}}(\left\| Z_{s, t}\right\|_\infty, \epsilon_2, k=\lceil e^{\epsilon_2/3} \rceil, U_t ) $ 
		\State Upload $ \left(Z_{s, t}/m_{\epsilon_1}, \mathcal{D}_s, U_{s, t} \right)  $ to the server
		\EndFor
		\State {\fontfamily{qcr}\selectfont //* Server Side Update *//}	
		\State Gradient update step $ \theta_{t+1} = \theta_t -  R^{-1}\frac{\eta}{B}\sum_{s \in S} Z^{\text{\textsf{dense}}}_{s, t} $, where 
		\State $ Z^{\text{\textsf{dense}}}_{s, t} $ is the $ d- $ dimensional zero vector with indices in $ \mathcal{D}_s $ replaced by $ Z_{s, t} $.
		\State Update gradient norm bound 
		\State $ U_{t+1} = \max(U_{t}, \max(\{U_{s, t}\}, s \in S_t)) $
		\EndFor
		\Return $ \theta_{T} $
	\end{algorithmic}
\end{algorithm}
\vspace{-1.5em}

Performing top-$ k $ selection with local privacy constraints could be done via two approaches. The first take is iteratively running the exponential mechanism \cite{dwork2014algorithmic} for $ k $ times, each time selecting a single index with gumble noise. The bottleneck of this take is that it requires sampling from a high dimensional distribution for $ k $ times, which is computationally heavy. Note that the privatization step is carried out on the client side device, which is usually assumed to be of limited computational power in FL settings \cite{kairouz2019advances}, thus using iterative exponential mechanism is not practical for FL scenarios. The second take is the noisy top-$ k $ algorithm \cite{ding2019free}, which generalizes the report noisy max mechanism in \cite{dwork2014algorithmic}. The algorithm requires adding Laplacian noise of scale $ 2 U k/\epsilon $ to each dimension of the gradient vector. In practice, $ k $ is typically chosen at the order of hundreds. Since most of the gradients are very small in magnitude, to ensure reasonable noise requires a high $ \epsilon $ budget to allocate for the selection step. This would significantly affect the overall privacy level. \par 
Motivated by algorithm designs in distributed learning like random masking \cite{konevcny2016federated} and Hogwild! \cite{hogwild}, we utilize a simple strategy, \emph{gradient subsampling}, by randomly sample $ \tilde{d} = \lfloor rd \rfloor $ dimensions, where $ r $ is the sampling proportion of gradient dimensions. To further improve training efficiency, we applied local accumulation techniques like momentum correction and factor masking similar to \cite{lin2018deep}. It is worth noting that low sampling ratio does \emph{not} necessarily improve model performance as the overall bias of the sparse approximation would become significant when only a few dimensions are selected. A more detailed study of this aspect is provided in Section \ref{subsample} of Appendix. 

\subsection{Improving quantization performance via randomized rotation}
As shown in \cite{agarwal2018cpsgd,suresh2017distributed}, for a $ d $-dimensional vector $ X $, the quantization error scales with $ \left(  \max_{j \in [d]} X^j - \min_{j \in [d]} X^j \right)^2  $. To reduce quantization error, an effective way is to preprocess the data via \emph{randomized rotation}, so as to make $ \max_{j \in [d]} X^j - \min_{j \in [d]} X^j  $ small in expectation. Note that we only need to rotate the fraction of gradient that are selected to get transmitted. \par 
To perform randomized rotation we need to assume public randomness between the server and the clients, under which we generate an orthogonal random matrix $ R \in \mathbb{R}^{\tilde{d} \times \tilde{d}} $, apply the linear transform to the individual gradients $ \tilde{X} = R X $ on the client side, and apply inverse transform on the server side $ \tilde{Z} = R^{-1} Z $. We adopt the method in \cite{agarwal2018cpsgd,suresh2017distributed} to generate $ R $ that supports fast matrix multiplication with $ O(d \log d) $ time and $ O(1) $ space. Specifically, we generate $ R = \frac{1}{\tilde{d}} HA $ where $ A $ is a random diagonal matrix with i.i.d. Rademacher entries (i.e. $ \mathbb{P}(A_{ii} = 1) =  \mathbb{P}(A_{ii} = -1) = 0.5, \forall i \in [\tilde{d}] $), and $ H $ is a Walsh-Hadamard matrix \cite{horadam2012hadamard}, defined recursively by the formula:
\begin{align}\label{hadamard}
	H(2^1) = 
	\left[ 
	\begin{array}{cc}
		1 & 1 \\
		1  & -1
	\end{array}
	\right] ,
	H(2^m) = 
	\left[ 
	\begin{array}{cc}
		H(2^{m-1}) & H(2^{m-1}) \\
		H(2^{m-1})  & -H(2^{m-1})
	\end{array}
	\right] ,
\end{align}
The randomized rotation operation is performed \emph{before} quantization and privatization, therefore we project the vector $ X $ to an $ \ell_2 $ ball of radius $ U $ before rotation. This would ensure the vector obtained after rotation has $ \ell_\infty $ norm bounded by $ U $ as well.

\subsection{Improving convergence via adaptive gradient norm shrinkage}
\label{sec:shrinkage}
$ \text{\textsf{PrivQuant}}_{K, \infty} $ quantizes its input $ X $ to $ \mathcal{B}^d $, regardless of the scale of $ X $ itself. This does not affect the unbiasedness of $ \text{\textsf{PrivQuant}}_{K, \infty} $ but may result in higher variance. Moreover in distributed optimization scenarios, a successful learning procedure implies that the magnitude of the gradients decreases during the learning process. Motivated by this phenomenon, we use an \emph{adaptive gradient norm shrinkage strategy} to iteratively refine the gradient norm upper bound $ U $ during the learning process, at the cost of an extra transmission of one real number for each round (both the server and selected clients). Specifically, we allocate a small fraction of the overall privacy budget to perform a $ \text{\textsf{ScalarDP}} $ \cite{bhowmick2018protection} mechanism over the $ \ell_\infty $ norm of the client's gradient vector and send it to the server together with privatized gradients and selected gradient dimensions. The server takes the maximum over the received gradient norm estimations as the estimation of gradient norm bound, and distributes it to the clients in the next round if the gradient norm bound decreases. 

\begin{figure}
	\centering
	\begin{tabular}{ccc}
		\includegraphics[width=45mm]{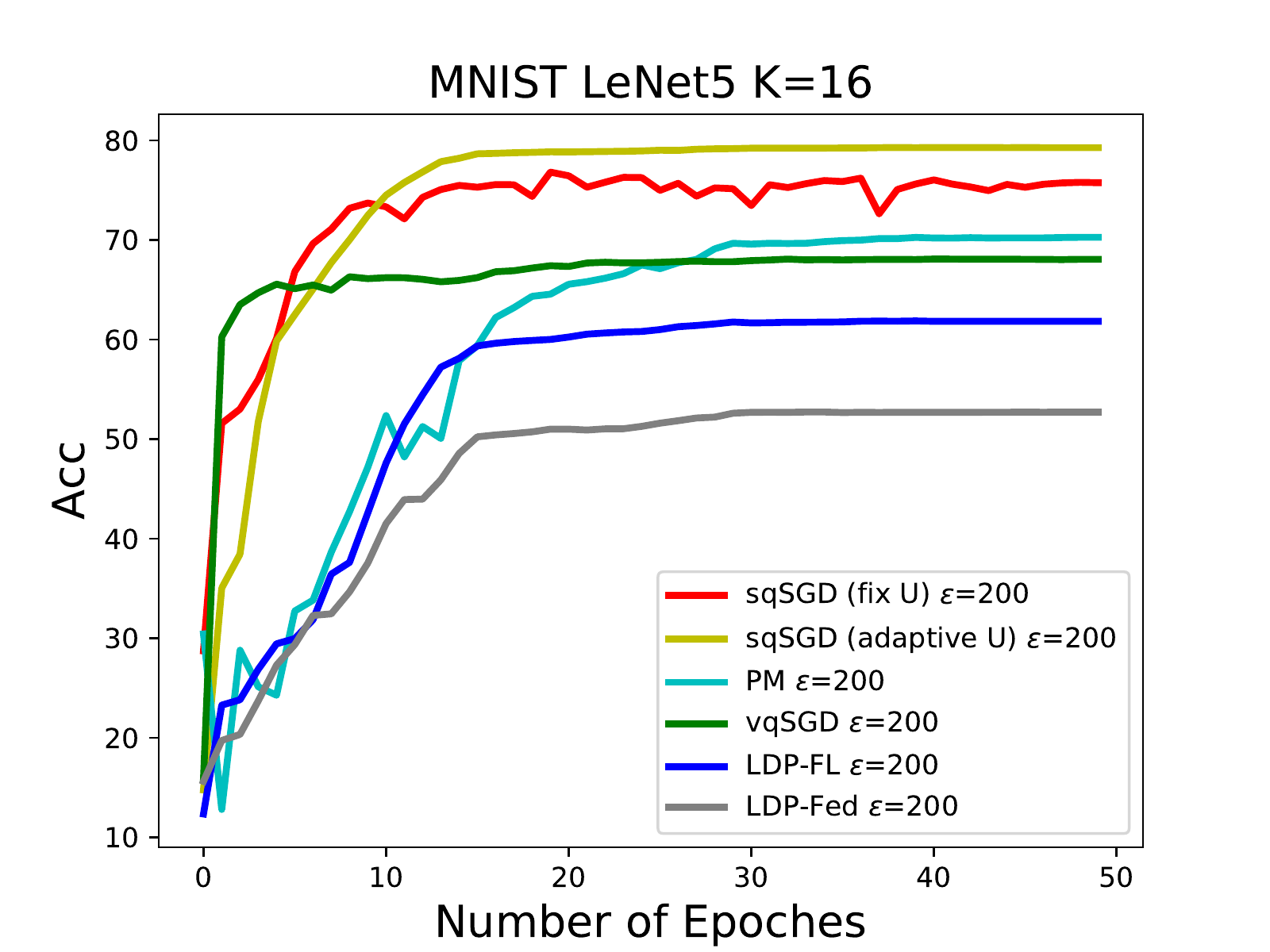} &
		\includegraphics[width=45mm]{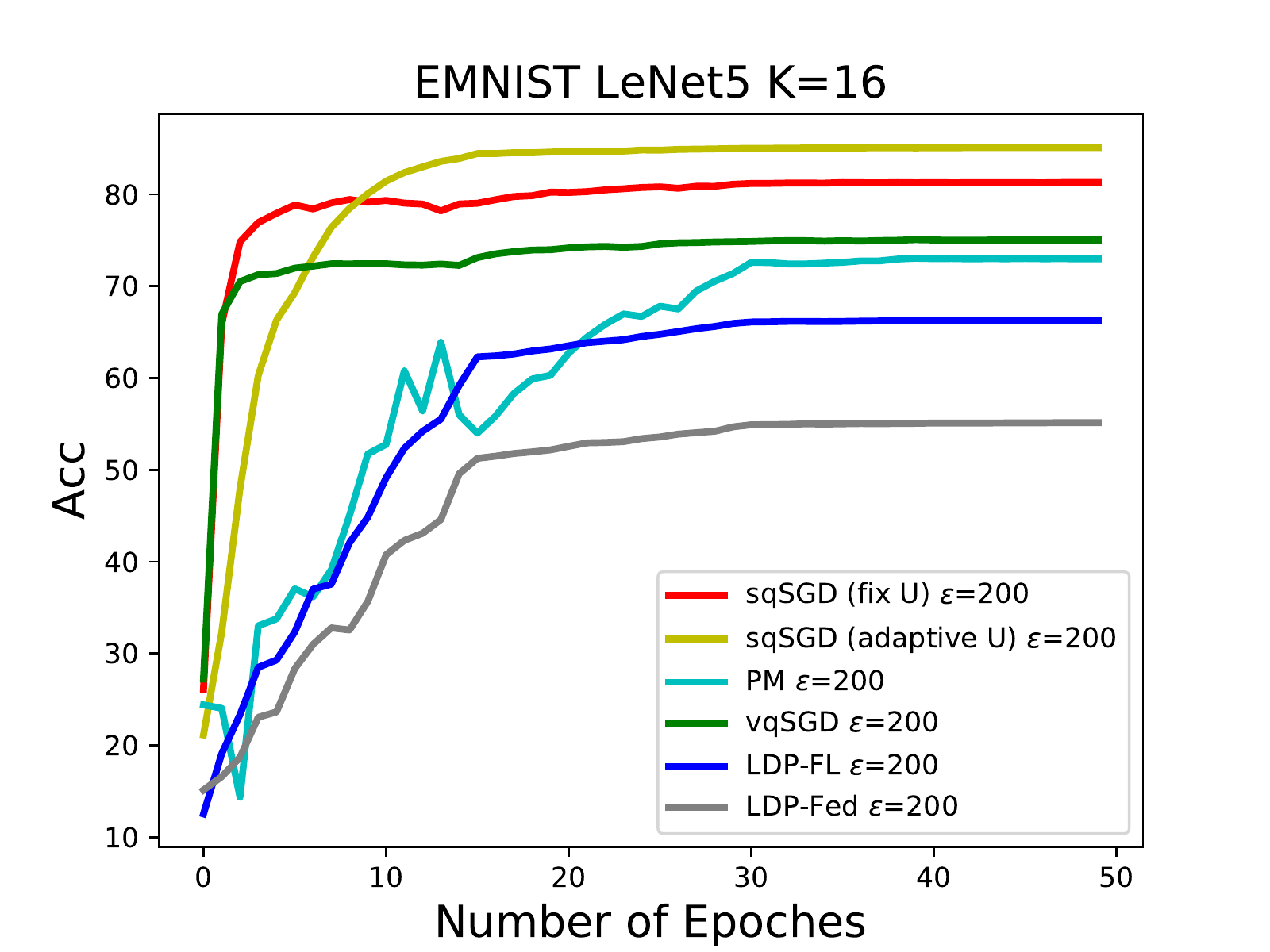} &
		\includegraphics[width=45mm]{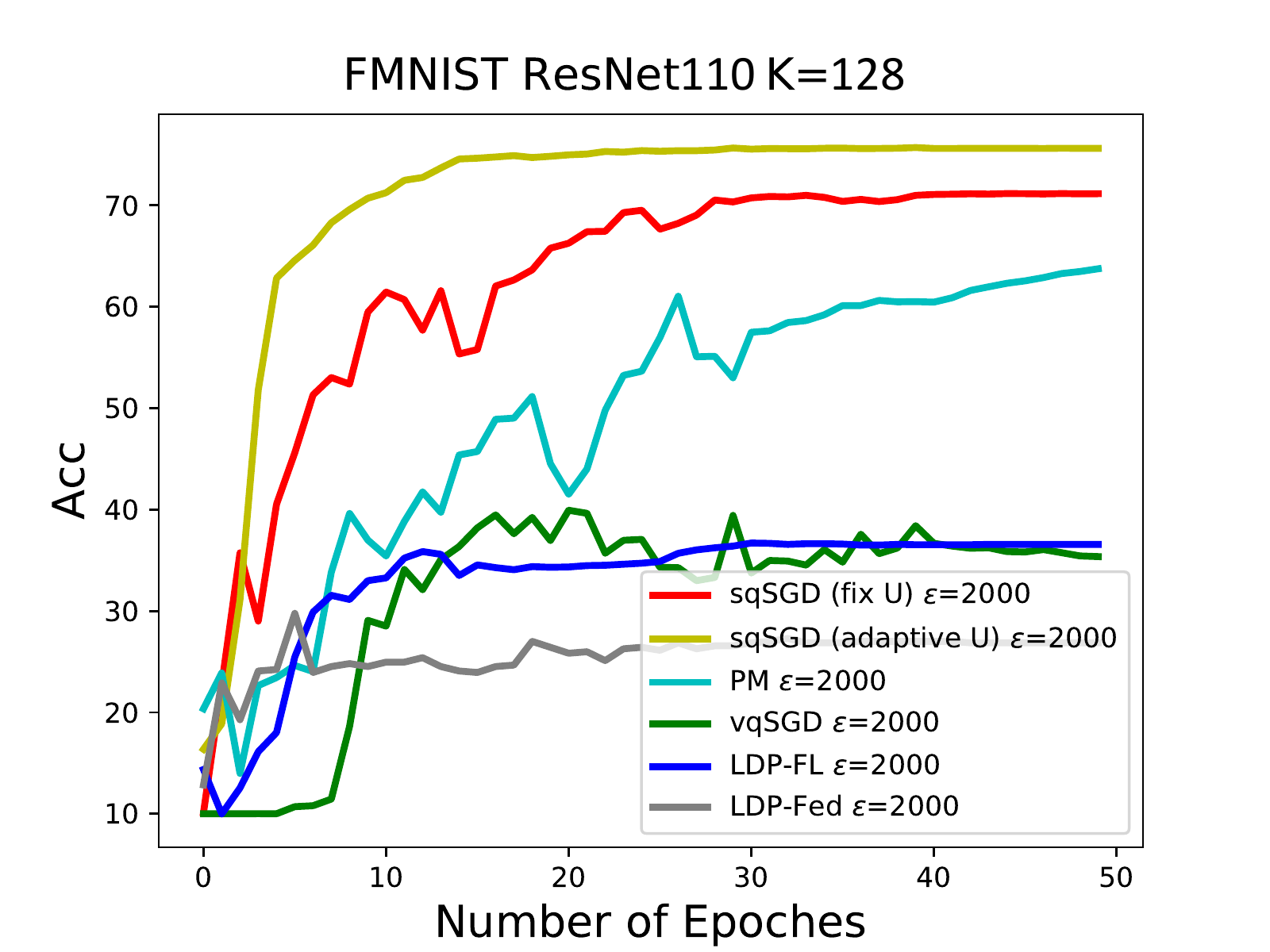} \\
		\includegraphics[width=45mm]{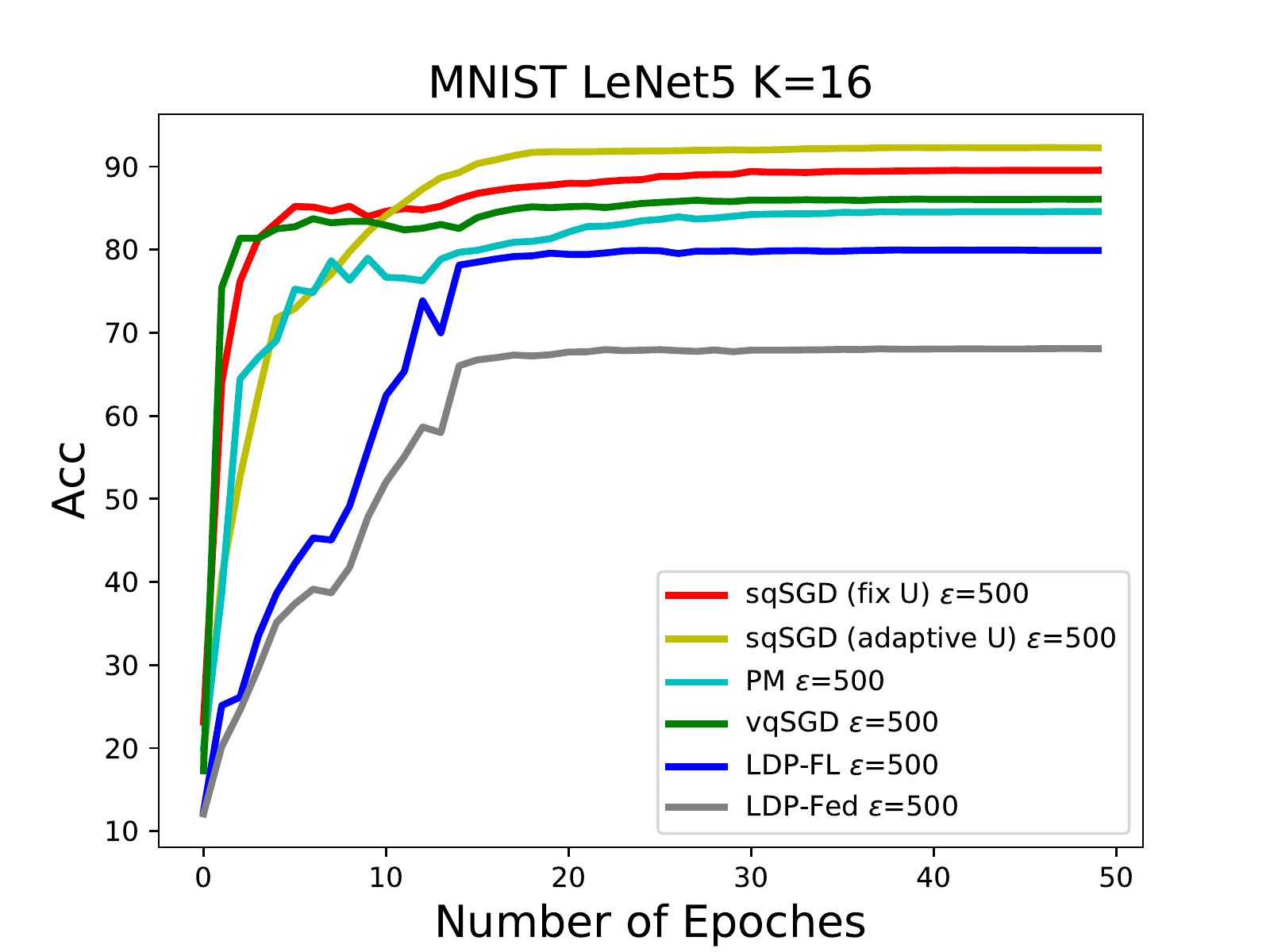} &
		\includegraphics[width=45mm]{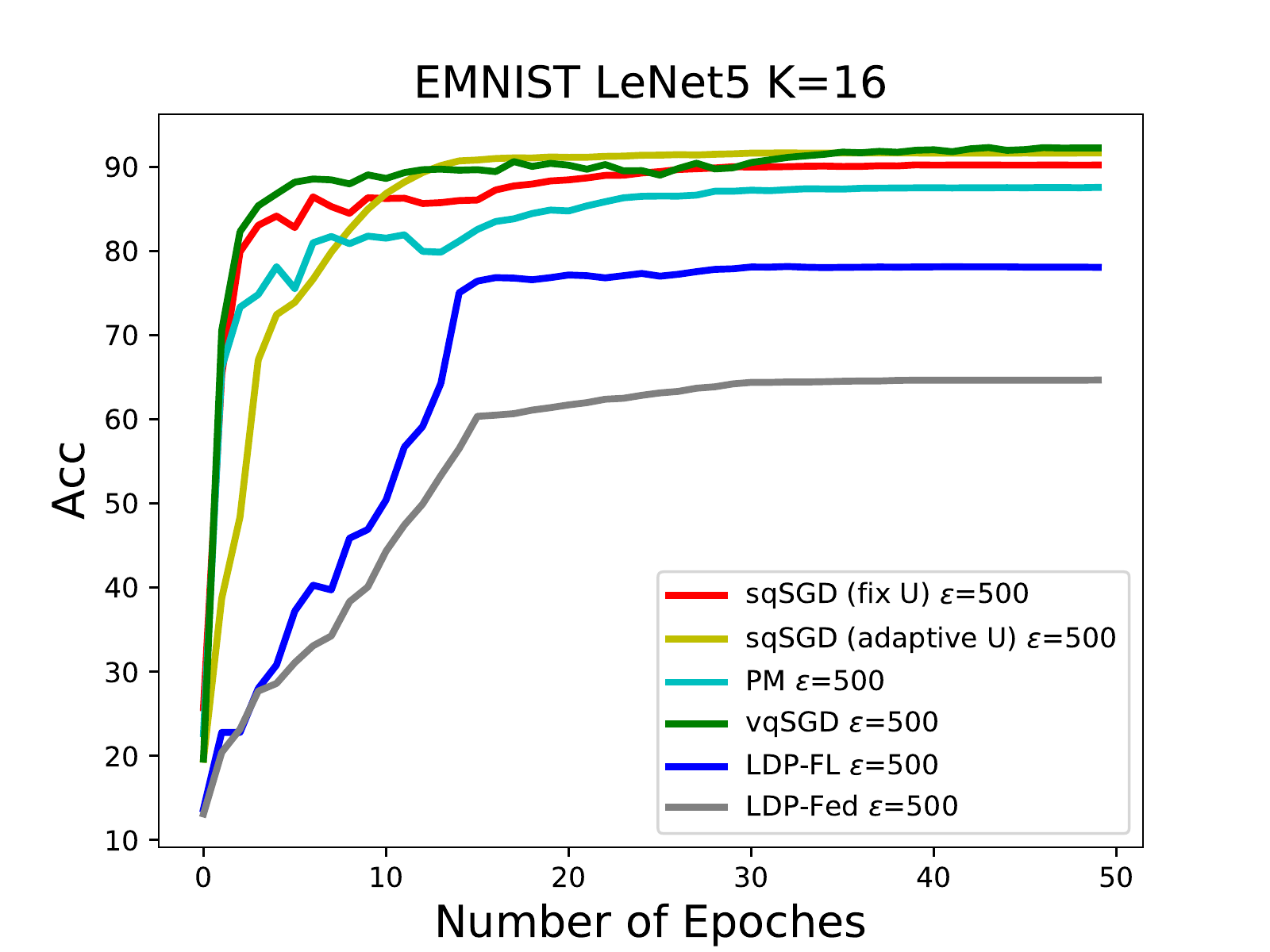} &
		\includegraphics[width=45mm]{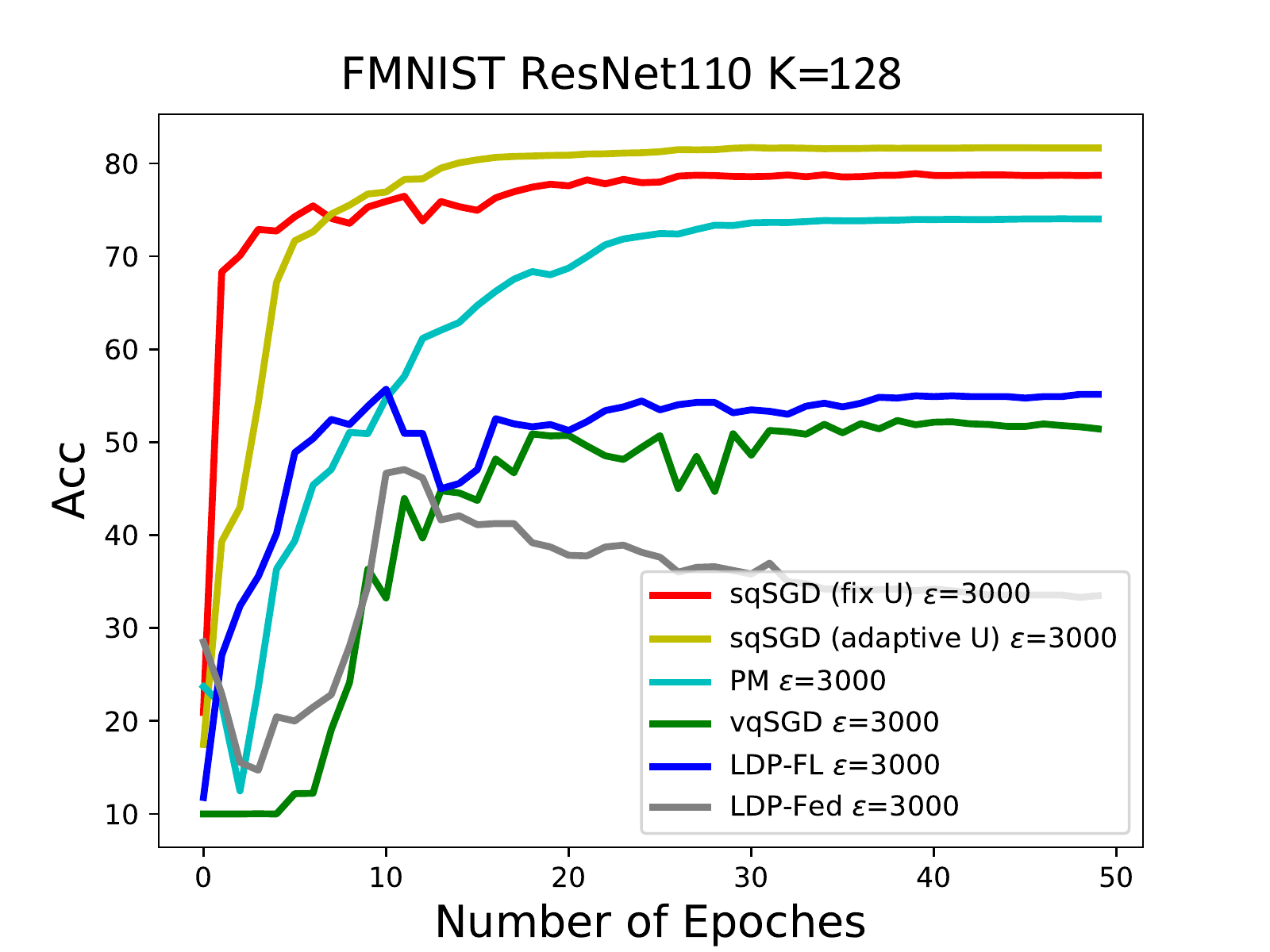} \\
	\end{tabular}
	\caption{Comparison of sqSGD against various baselines}
	\label{fig: 5}
\end{figure}

The final selective SGD updating process is summarized in Algorithm \ref{alg: privSQ}. The randomized rotation and sparsification strategy have no effect on privacy level, hence the local update within a single round is $ \epsilon$-LDP. At the same time, communication cost was further reduced from $ O(d\log_2 K) $ bits to $ O(\tilde{d}\log_2 K) $ bits per client.

\section{Experiments} \label{experiments}
In this section, we apply sqSGD to the simulated FL environments, constructed via standard datasets MNIST \cite{LeNet},  EMNSIT \cite{cohen2017emnist}, and Fashion MNIST \cite{xiao2017/online} which we abbreviate as FMNIST hereafter. All three datasets are of equal (training) sample size, therefore we randomly partition each dataset into $ 10 $ equally sized blocks to simulate a \emph{cross-silo} FL environment consisting of a central server and $ 10 $ clients. We train on MNIST and EMNIST datasets using the  LeNet-5 architecture \cite{LeNet} ($ 119,850 $ parameters). We also train a larger network under the ResNet-110 architecture ($ 1,722,224 $ parameters) on the FMNIST dataset.  Due to the limited space, we provide the experimental details and ablation studies of sqSGD in Sec. \ref{setting} and \ref{ablation} of Appendix.\par

\subsection{Baseline comparisons} To validate the efficacy of our sqSGD, we compare with the following four baselines. 
\begin{description}
	\item[Piecewise mechanism (PM) \cite{wang2019collecting}] PM is designed for mean estimation under local privacy model that improves per-coordinate accuracy upon the $ \textsf{PrivUnit}_\infty $ algorithm in \cite{bhowmick2018protection} with $ \kappa=0 $. Although PM communicates real values, the multi-dimensional version of PM \cite[Algorithm 4]{wang2019collecting} samples a fraction of coordinates for perturbation hence we may view PM practically as a limited-communication mean estimation algorithm. For large $ d $ settings, we choose reasonably large $ \epsilon $ values such that PM communicates raw perturbed gradients in $ \lfloor \epsilon/2.5\rfloor $ dimensions.
	\item[LDP-Fed \cite{truex2020ldp}] LDP-Fed is a FL framework that adopts the condensed LDP (CLDP) protocol \cite{cldp}, which is a generalization of the LDP that allows the privacy loss random variable to depend on some distance $ d(\cdot, \cdot) $ between its input data. In our experiments we pick $ d $ to be the $ \ell_1 $ metric, and the privatization scheme is chosen as the exponential-type mechanism used in \cite{truex2020ldp}. We choose the clipping parameter so that communication cost is the same across all baselines. We use the converted privacy budget in LDP model according to the formula in \cite{cldp}. 
	\item[LDP-FL \cite{sun2020ldp}] LDP-FL adopts a privatization scheme similar to \cite{wang2019collecting}, and presumes the existence of a trusted third-party that guarantees client anonymity. In our experiments we do not rely on this third-party device, and only test the efficiency of LDP-FL's privatization algorithm. Hyperparameters are chosen so that communication costs are equal across all baselines.
	\item[vqSGD \cite{g2019vqsgd}] vqSGD uses an communication-optimal distributed mean estimation algorithm that adopts vector quantization strategies. The per client communication cost of vqSGD is $ O(\log d) $ for mean estimation task. The algorithm has trivial extensions to locally private settings using randomized response strategies \cite{g2019vqsgd}. As discussed by the authors, for distributed optimization settings with high dimensional models, variance reduction techniques are necessary, specifically one data point is communicated $ b $ times and server uses the average of  $ b $ quantized gradients for aggregation. This makes the communication cost $ O(b\log d) $. 
\end{description}
Note that we compare all the approaches under the same privacy budget and communication cost. For sqSGD we use $ K = 16 $ for LeNet and $ K = 128 $ for ResNet, and we adjust the sampling ratio in sqSGD such that communication costs remain equal among all methods. We present both sqSGD with fixed gradient norm bound at $ U = 10 $ and with the adaptive norm shrinking strategy as described in Sec~\ref{sec:shrinkage}. The comparisons are presented in Figure \ref{fig: 5}, which shows that sqSGD with fixed $ U $ consistently outperforms PM, LDP-Fed and LDP-FL by a significant margin across all tasks under various privacy levels. vqSGD obtains comparable performance with sqSGD in relatively smaller models and high-privacy regime (Note that vqSGD offers much worse privacy protections in this setting), but breaks down when models get larger. Moreover, utilizing the adaptive norm shrinking strategy increases final model accuracy by over $ 2\% $ on average, meanwhile, training stability is also significantly improved. Overall, in high-dimensional settings, sqSGD provides a much better optimization performance.

\section{Related Work}
\textbf{Differentially private federated learning.} FL with differential privacy constraints is an active area of research. The majority of previous works adopt the distributed optimization setup with ERM objective using SGD based learning algorithms, while differing in their privacy models. Most private FL frameworks use central DP model \cite{PrivDP,brendan2018learning,feddp}. \cite{NEURIPS2018_7221e5c8} combines zero-concentrated differential privacy mechanisms with secure multi-party composition methods. \cite{wang2021dpfeddifferentially} adopts R\'{e}nyi differential privacy model. The local differential privacy model offers stronger privacy protection, and has reasonable performance under low dimensional learning scenarios \cite{duchi2018minimax,wang2019collecting}. Scaling locally private FL to high-dimensional models require additional assumptions on the problem. \cite{bhowmick2018protection} restricts the prior knowledge of the adversary to allow high $\epsilon$ training with reasonable protection guarantee against reconstruction. \cite{sun2020ldp} proposes to use a secure data processing procedure called \emph{splitting and shuffling} to bypass the curse of dimensionality. \par 
\textbf{Communication-efficient federated learning.} Since its introduction in \cite{konevcny2016federated}, communication efficiency has been one of the central topics of FL \cite{konevcny2016federated,kairouz2019advances}. \cite{konevcny2016federated} described two kinds of general approaches for improving communication efficiency: structured updates like randomized masking or gradient dropping, and sketched updates like quantization methods. Below we review these two approaches separately. 
\begin{description}
	\item[Gradient sparsification.] Gradient dropping methods threshold gradients based on either fixed single threshold \cite{aji2017sparse} or adaptively chosen thresholds \cite{chen2017adacomp} 
	which can greatly reduce communication overhead. DGC \cite{lin2018deep} adopted more elegant refinements like accumulation correction and factor masking to achieve even higher compression ratios. FedSel \cite{liu2020fedsel} performed privatized gradient sparsification via adopting randomized response techniques to select the most significant gradient dimension. 
	\item[Gradient quantization.] Quantization is another important type of mechanisms that significantly reduces communication cost of distributed learning processes. As far as we have noticed, there are two types of quantization schemes that are applicable to distributed learning scenarios with privacy constraints. The first type is \emph{coordinate-wise quantization}, where each coordinate of a $ d- $dimensional vector is separately quantized to a given range represented by $ b $ bits. The communication cost of this type of algorithms is $ O(db) $. Methods like QSGD \cite{qsgd} or TernGrad \cite{terngrad} are able to achieve little performance degradation on large neural architectures using less than $ 4 $ quantization bits. Privatized versions of quantized SGD are recently proposed, cpSGD \cite{agarwal2018cpsgd} adopts the distributed DP model to perform private distributed learning. The second type is \emph{vector quantization} that directly quantizes the input vector to a pre-selected set of $ d- $dimensional vectors. vqSGD \cite{g2019vqsgd} with its cross polytope scheme requires only $ O(\log d) $ bits of communication. SQKR \cite{chen2020breaking} is another powerful quantization scheme that achieves minimax optimality under both communication and privacy constraints.
\end{description}

\section{Conclusion}
We studied privacy-preserving federated learning under the local differential privacy model. To achieve both communication efficiency and high-dimensional compatibility, we proposed a gradient-based learning framework sqSGD that is based on a novel private multivariate mean estimation scheme with controllable quantization levels. We applied gradient subsampling and randomized rotation to reduce estimation error of the base mechanism that exploits the specific structure of federated learning with large modern neural architectures. Finally, 
on benchmark datasets, our sqSGD is capable of training large models with random initializations, surpassing baseline algorithms by a significant margin.

{\small
\bibliographystyle{splncs04}
\bibliography{sample-base}

\begin{thebibliography}{10}
\providecommand{\url}[1]{\texttt{#1}}
\providecommand{\urlprefix}{URL }
\providecommand{\doi}[1]{https://doi.org/#1}

\bibitem{PrivDP}
Abadi, M., Chu, A., Goodfellow, I., McMahan, e.a.: Deep learning with
  differential privacy. In: Proceedings of the 2016 ACM SIGSAC Conference on
  Computer and Communications Security. CCS '16, Association for Computing
  Machinery, New York, NY, USA (2016)

\bibitem{agarwal2018cpsgd}
Agarwal, N., Suresh, A.T., Yu, F.X.X., Kumar, S., McMahan, B.: cpsgd:
  Communication-efficient and differentially-private distributed sgd. In:
  Advances in Neural Information Processing Systems. pp. 7564--7575 (2018)

\bibitem{aji2017sparse}
Aji, A.F., Heafield, K.: Sparse communication for distributed gradient descent.
  In: Proceedings of the 2017 Conference on Empirical Methods in Natural
  Language Processing. pp. 440--445 (2017)

\bibitem{qsgd}
Alistarh, D., Grubic, D., Li, J., Tomioka, R., Vojnovic, M.: Qsgd:
  Communication-efficient sgd via gradient quantization and encoding. In:
  Advances in Neural Information Processing Systems 30, pp. 1709--1720 (2017)

\bibitem{FL}
Apple.: Private federated learning (neurips 2019 expo talk abstract) (2019)

\bibitem{bhowmick2018protection}
Bhowmick, A., Duchi, J., Freudiger, J., Kapoor, G., Rogers, R.: Protection
  against reconstruction and its applications in private federated learning.
  arXiv preprint arXiv:1812.00984  (2018)

\bibitem{chen2017adacomp}
Chen, C.Y., Choi, J., Brand, D., Agrawal, A., Zhang, W., Gopalakrishnan, K.:
  Adacomp : Adaptive residual gradient compression for data-parallel
  distributed training (2017)

\bibitem{chen2020breaking}
Chen, W.N., Kairouz, P., {\"O}zg{\"u}r, A.: Breaking the
  communication-privacy-accuracy trilemma. arXiv preprint arXiv:2007.11707
  (2020)

\bibitem{cohen2017emnist}
Cohen, G., Afshar, S., Tapson, J., van Schaik, A.: Emnist: an extension of
  mnist to handwritten letters (2017)

\bibitem{ding2019free}
Ding, Z., Wang, Y., Zhang, D., Kifer, D.: Free gap information from the
  differentially private sparse vector and noisy max mechanisms (2019)

\bibitem{duchi2019lower}
Duchi, J., Rogers, R.: Lower bounds for locally private estimation via
  communication complexity. In: Conference on Learning Theory. pp. 1161--1191
  (2019)

\bibitem{duchi2018minimax}
Duchi, J.C., Jordan, M.I., Wainwright, M.J.: Minimax optimal procedures for
  locally private estimation. Journal of the American Statistical Association
  \textbf{113}(521),  182--201 (2018)

\bibitem{duchi2018right}
Duchi, J.C., Ruan, F.: The right complexity measure in locally private
  estimation: It is not the fisher information. arXiv preprint arXiv:1806.05756
   (2018)

\bibitem{dwork2014algorithmic}
Dwork, C., Roth, A., et~al.: The algorithmic foundations of differential
  privacy. Foundations and Trends in Theoretical Computer Science
  \textbf{9}(3-4),  211--407 (2014)

\bibitem{g2019vqsgd}
Gandikota, V., Kane, D., Maity, R.K., Mazumdar, A.: vqsgd: Vector quantized
  stochastic gradient descent (2019)

\bibitem{cldp}
{Gursoy}, M.E., {Tamersoy}, A., {Truex}, S., {Wei}, W., {Liu}, L.: Secure and
  utility-aware data collection with condensed local differential privacy. IEEE
  Transactions on Dependable and Secure Computing pp.~1--1 (2019).
  \doi{10.1109/TDSC.2019.2949041}

\bibitem{hard2019federated}
Hard, A., Rao, K., Mathews, R., Ramaswamy, S., Beaufays, F., Augenstein, S.,
  Eichner, H., Kiddon, C., Ramage, D.: Federated learning for mobile keyboard
  prediction (2019)

\bibitem{he2016deep}
He, K., Zhang, X., Ren, S., Sun, J.: Deep residual learning for image
  recognition. In: Proceedings of the IEEE conference on computer vision and
  pattern recognition. pp. 770--778 (2016)

\bibitem{horadam2012hadamard}
Horadam, K.J.: Hadamard matrices and their applications. Princeton university
  press (2012)

\bibitem{NEURIPS2018_7221e5c8}
Jayaraman, B., Wang, L., Evans, D., Gu, Q.: Distributed learning without
  distress: Privacy-preserving empirical risk minimization. In: Advances in
  Neural Information Processing Systems. pp. 6343--6354 (2018)

\bibitem{kairouz2019advances}
Kairouz, P., McMahan, H.B., Avent, B., Bellet, A., Bennis, M., Bhagoji, A.N.,
  et~al: Advances and open problems in federated learning (2019)

\bibitem{kasiviswanathan2011can}
Kasiviswanathan, S.P., Lee, H.K., Nissim, K., Raskhodnikova, S., Smith, A.:
  What can we learn privately? SIAM Journal on Computing  \textbf{40}(3),
  793--826 (2011)

\bibitem{konevcny2016federated}
Kone{\v{c}}n{\`y}, J., McMahan, H.B., Yu, F.X., Richt{\'a}rik, P., Suresh,
  A.T., Bacon, D.: Federated learning: Strategies for improving communication
  efficiency. arXiv preprint arXiv:1610.05492  (2016)

\bibitem{LeNet}
{Lecun}, Y., {Bottou}, L., {Bengio}, Y., {Haffner}, P.: Gradient-based learning
  applied to document recognition. Proceedings of the IEEE  \textbf{86}(11),
  2278--2324 (1998)

\bibitem{leroy2019fed}
{Leroy}, D., {Coucke}, A., {Lavril}, T., {Gisselbrecht}, T., {Dureau}, J.:
  Federated learning for keyword spotting. In: ICASSP 2019 - 2019 IEEE
  International Conference on Acoustics, Speech and Signal Processing (ICASSP).
  pp. 6341--6345 (2019)

\bibitem{lin2018deep}
Lin, Y., Han, S., Mao, H., Wang, Y., Dally, B.: Deep gradient compression:
  Reducing the communication bandwidth for distributed training. In:
  International Conference on Learning Representations (2018),
  \url{https://openreview.net/forum?id=SkhQHMW0W}

\bibitem{liu2020fedsel}
Liu, R., Cao, Y., Yoshikawa, M., Chen, H.: Fedsel: Federated sgd under local
  differential privacy with top-k dimension selection (2020)

\bibitem{lyu2020privacy}
Lyu, L., Yu, H., Ma, X., Sun, L., Zhao, J., Yang, Q., Yu, P.S.: Privacy and
  robustness in federated learning: Attacks and defenses. arXiv preprint
  arXiv:2012.06337  (2020)

\bibitem{lyu2020threats}
Lyu, L., Yu, H., Zhao, J., Yang, Q.: Threats to federated learning. In:
  Federated Learning, pp. 3--16. Springer (2020)

\bibitem{mcmahan2017communication}
McMahan, B., Moore, E., Ramage, D., Hampson, S., y~Arcas, B.A.:
  Communication-efficient learning of deep networks from decentralized data.
  In: Artificial Intelligence and Statistics. pp. 1273--1282. PMLR (2017)

\bibitem{brendan2018learning}
McMahan, H.B., Ramage, D., Talwar, K., Zhang, L.: Learning differentially
  private recurrent language models. In: International Conference on Learning
  Representations (2018), \url{https://openreview.net/forum?id=BJ0hF1Z0b}

\bibitem{melis2018inference}
Melis, L., Song, C., De~Cristofaro, E., Shmatikov, V.: Inference attacks
  against collaborative learning. arXiv preprint arXiv:1805.04049  \textbf{13}
  (2018)

\bibitem{hogwild}
Niu, F., Recht, B., Re, C., Wright, S.J.: Hogwild! a lock-free approach to
  parallelizing stochastic gradient descent. In: Proceedings of the 24th
  International Conference on Neural Information Processing Systems (2011)

\bibitem{sun2020ldp}
Sun, L., Qian, J., Chen, X., Yu, P.S.: Ldp-fl: Practical private aggregation in
  federated learning with local differential privacy. arXiv preprint
  arXiv:2007.15789  (2020)

\bibitem{suresh2017distributed}
Suresh, A.T., Felix, X.Y., Kumar, S., McMahan, H.B.: Distributed mean
  estimation with limited communication. In: International Conference on
  Machine Learning. pp. 3329--3337 (2017)

\bibitem{truex2020ldp}
Truex, S., Liu, L., Chow, K.H., Gursoy, M.E., Wei, W.: Ldp-fed: Federated
  learning with local differential privacy. In: Proceedings of the Third ACM
  International Workshop on Edge Systems, Analytics and Networking. pp. 61--66
  (2020)

\bibitem{wang2021dpfeddifferentially}
Wang, L., Jia, R., Song, D.: D2p-fed:differentially private federated learning
  with efficient communication (2021),
  \url{https://openreview.net/forum?id=wC99I7uIFe}

\bibitem{wang2019collecting}
Wang, N., Xiao, X., Yang, Y., Zhao, J., Hui, S.C., Shin, H., Shin, J., Yu, G.:
  Collecting and analyzing multidimensional data with local differential
  privacy. In: 2019 IEEE 35th International Conference on Data Engineering
  (ICDE). pp. 638--649. IEEE (2019)

\bibitem{feddp}
{Wei}, K., {Li}, J., {Ding}, M., {Ma}, C., {Yang}, H.H., {Farokhi}, F., {Jin},
  S., {Quek}, T.Q.S., {Vincent Poor}, H.: Federated learning with differential
  privacy: Algorithms and performance analysis. IEEE Transactions on
  Information Forensics and Security  \textbf{15},  3454--3469 (2020)

\bibitem{terngrad}
Wen, W., Xu, C., Yan, F., Wu, C., Wang, Y., Chen, Y., Li, H.: Terngrad: Ternary
  gradients to reduce communication in distributed deep learning. In: Guyon,
  I., Luxburg, U.V., Bengio, S., Wallach, H., Fergus, R., Vishwanathan, S.,
  Garnett, R. (eds.) Advances in Neural Information Processing Systems 30, pp.
  1509--1519. Curran Associates, Inc. (2017)

\bibitem{xiao2017/online}
Xiao, H., Rasul, K., Vollgraf, R.: Fashion-mnist: a novel image dataset for
  benchmarking machine learning algorithms (2017)

\end{thebibliography}
}
\appendix
\newpage

\section{Proofs}\label{appendix: A1}
\begin{proof}[Proof of theorem 1]
	Let $u \in \mathcal{B}^d$ and
	$U \sim \text{\textsf{Unif}}(\mathcal{B}^d)$.  The vector $V \in \mathcal{B}^d$
	sampled as in Algorithm \ref{alg: privQ}, has p.m.f.
	\begin{equation*}
		p(v \mid u)
		\propto \begin{cases} 1 / \P(\mathcal{M}(U, u) > \kappa) & \mbox{if~}
			\mathcal{M}(v, u) > \kappa \\
			1 / \P(\mathcal{M}(U, u) < \kappa) & \mbox{if} \mathcal{M}(v, u) \leq \kappa.
		\end{cases}
	\end{equation*}
	The event that $\mathcal{M}(U, u) = \kappa$ when $\frac{d+\kappa+1}{2} \in \mathbb{Z}$
	implies that $U$ and $u$ match in exactly $\frac{d+\kappa+1}{2}$
	coordinates; the number of such matches is
	${d}\choose{(d+\kappa+1)/2}$. Computing the binomial sum, we have
	\begin{align}\label{probdef}
		\begin{split}
			\P(\mathcal{M}(U, u)  > \kappa) &=  \frac{1}{K^d}
			\sum_{\ell = \lceil \frac{d+\kappa+1}{2} \rceil}^d
			{{d}\choose{\ell}} (K - 1)^{(d - \ell)} \\
			\quad
			\P(\mathcal{M}(U, u)  \leq \kappa) &=  \frac{1}{K^d}
			\sum_{\ell = 0 }^{\lceil\frac{d+\kappa+1}{2}\rceil -1} {{d}\choose{\ell}} (K - 1)^{(d - \ell)}
		\end{split}
	\end{align}
	Now we show unbiasedness via showing $\E[V \mid u = u] = m \cdot u$. We have 
	\begin{align*}
		\E&[V \mid u = u] \\
		&= p \E[U \mid \mathcal{M}(U, u) > \kappa]  +
		(1 - p) \E[U \mid \mathcal{M}(U, u) \leq \kappa]
	\end{align*}
	By rotational symmetry, it suffices to show:
	\begin{align*}
		\E&[V^1 \mid u = u] \\
		&= p \underbrace{\E[U^1 \mid \mathcal{M}(U, u) > \kappa]}_{\Upsilon_1} +
		(1 - p) \underbrace{\E[U^1 \mid \mathcal{M}(U, u) \leq \kappa]}_{\Upsilon_2}
	\end{align*}
	For $ \Upsilon_1 $, we have:
	\begin{align*}
		\Upsilon_1 &= \dfrac{1}{K^d  \P(\mathcal{M}(U, u) > \kappa) } \times\sum_{\ell =  \tau}^d \left( u^1 {{d}\choose{\ell}} (K - 1)^{d - \ell} \right. \\
		& \left. \quad - \sum_{w \in \mathcal{B}\backslash u^1} w {{d}\choose{\ell}}(K - 1)^{d - \ell - 1} \right) \\
		& =  \dfrac{u^1}{K^d  \P(\mathcal{M}(U, u) > \kappa) } \times {{d-1}\choose{ \tau-1}} (K - 1)^{(d - \tau)}
	\end{align*}
	Where in the second equality we used the fact that $ \sum_{w \in \mathcal{B}} w = 0 $. Similar calculations yield:
	\begin{align*}
		\Upsilon_2 = - \dfrac{u^1}{K^d  \P(\mathcal{M}(U, u) \le \kappa) } \times {{d-1}\choose{ \tau-1}} (K - 1)^{(d - \tau)}
	\end{align*}
	Combining the preceding display with \eqref{probdef}, we have:
	\begin{align*}
		\E&[V^1 \mid u = u]  \\
		&= \left( p\frac{{d-1 \choose{ \tau - 1} }(K - 1)^{d - \tau} }{\sum_{\ell = \tau}^d {{d}\choose{\ell}} (K - 1)^{d - l}}
		- (1 - p) \frac{ {{d-1}\choose{d \tau -1 }} (K - 1)^{d - \tau} }{\sum_{\ell = 0}^{\tau -1 } {{d}\choose{\ell}} (K - 1)^{d-l}} \right) \\
		&\qquad \times u^1 \\
		&= m u^1
	\end{align*}
	Next we show privacy guarantee. 
	\begin{description}
		\item[LDP guarantee] As $\P(\mathcal{M}(U, u)  > \kappa)$ is decreasing in $\kappa$ for any $u, u' \in
		\mathcal{B}^d$ and $v \in \mathcal{B}^d$ we have
		\begin{align*}
			\frac{p(v \mid u) }{p(v \mid u')} &\leq \frac{p}{1 - p}
			\cdot \frac{\P(\mathcal{M}(U, u^\prime)  \leq \kappa)}{\P(\mathcal{M}(U, u) > \kappa)}\\
			&= \frac{p}{1 - p}
			\times
			\frac{\sum_{\ell =0}^{ \tau-1 }{{d}\choose{\ell}} (K - 1)^{(d - \ell)}}{
				\sum_{\ell = \tau }^d {{d}\choose{\ell}} (K - 1)^{(d - \ell)}}
		\end{align*}
		The result follows by relation \eqref{eq:sufficient_kappaGEN}
		\item[Reconstruction protection guarantee] the result follows by \cite[Proposition 1]{bhowmick2018protection} and the LDP guarantee.
	\end{description}
	
\end{proof}

\section{Experimental Settings}\label{setting}
Since the main goal of the empirical study is not to achieve state-of-the-art performance, we adopt the following common strategies without further tuning across all experiments: in each round, we randomly sample a single batch with batch size $ 32 $ for each client and compute a single gradient update step. Without gradient subsampling, this would be equivalent to a stochastic gradient descent update over the entire data with a batch size of $ 320 $. The training duration is measured via number of epochs ($ 1 $ epoch $ = 188 $ rounds). We use the same configuration of learning rate $ \eta = 0.001 $ and local correction factors $ \alpha=1.0, \beta=1.0 $ across all experiments. We set the initial $ \ell_2 $ norm bound of the gradients to be $ U = 10 $. Finally, for a given privacy budget $ \epsilon $, we set $ \epsilon_2 = 10 $ if the adaptive gradient norm bound shrinkage strategy is adopted. For the remaining budget $ \epsilon_1 = \epsilon - \epsilon_2 $ we use the following method to determine the value of $ (\kappa_{\epsilon_1}, p_{\epsilon_1}) $: We set $ \kappa_{\epsilon_1} $ to be the largest integer such that the value of the left hand side of \eqref{eq:sufficient_kappaGEN} is smaller than or equal to $ e^{0.9\epsilon_1} $, and $ p_{\epsilon_1}= \frac{e^{0.1\epsilon_1}}{1 + e^{0.1\epsilon_1}} $. \par

\textbf{Choice of privacy budget $ \epsilon $.} Across all the experiments we choose $ \epsilon = O(\sqrt{d}) $. As the models used in our experiments are all of high dimension, the choice of high $\epsilon$ will basically provide little protection against inference attacks. However, with respect to reconstruction attack specified in Section \ref{sec: meth} and the protection level \eqref{eqn: protection_reconstruction}, such choice still provides reasonable protection against reconstruction attacks if the adversary's goal is to evaluate more than $ O(\sqrt{d}) $ of the entries in the private data. \par 
\begin{figure*}
	\centering
	\begin{tabular}{ccc}
		\includegraphics[width=45mm]{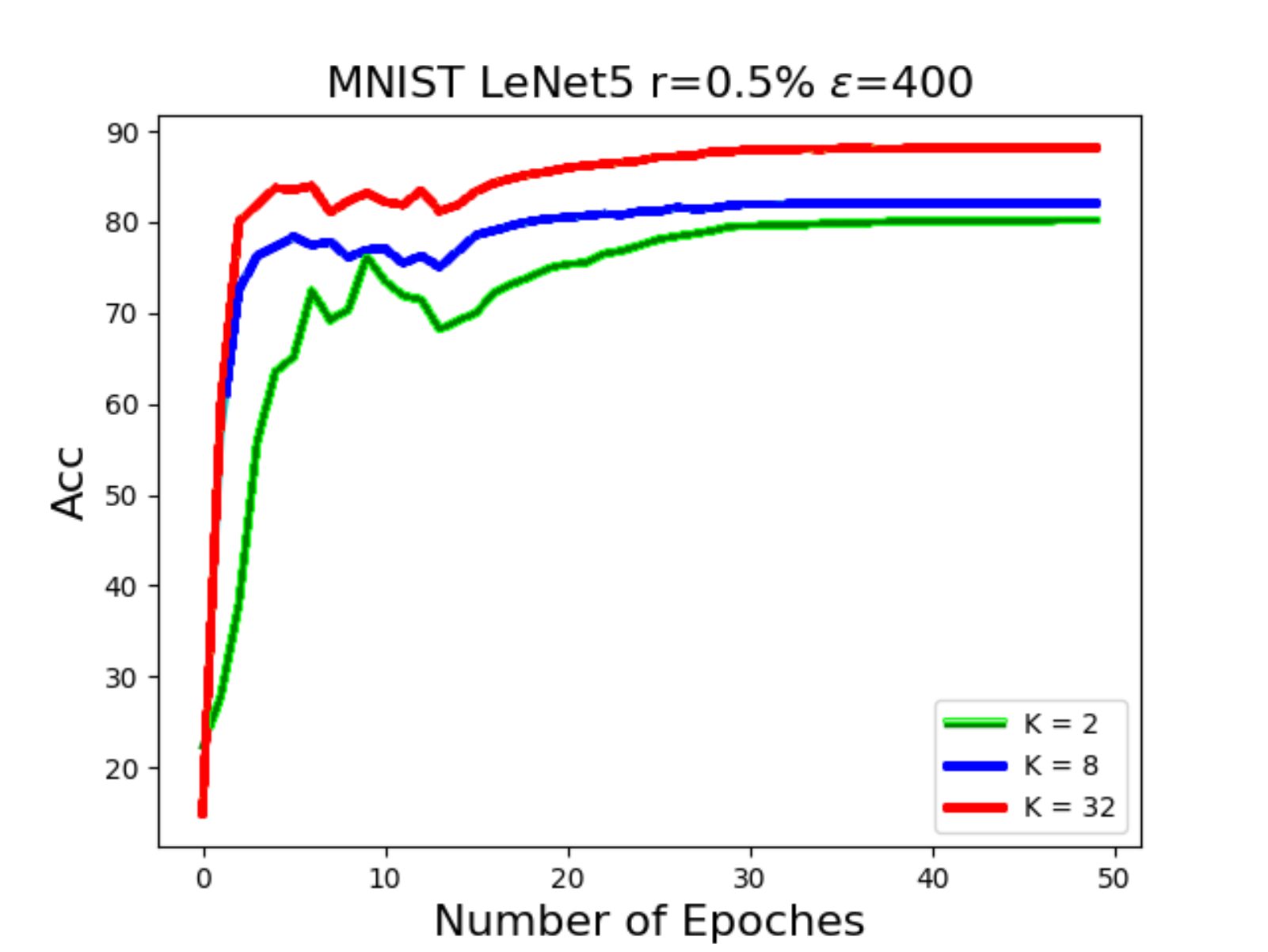}
		\includegraphics[width=45mm]{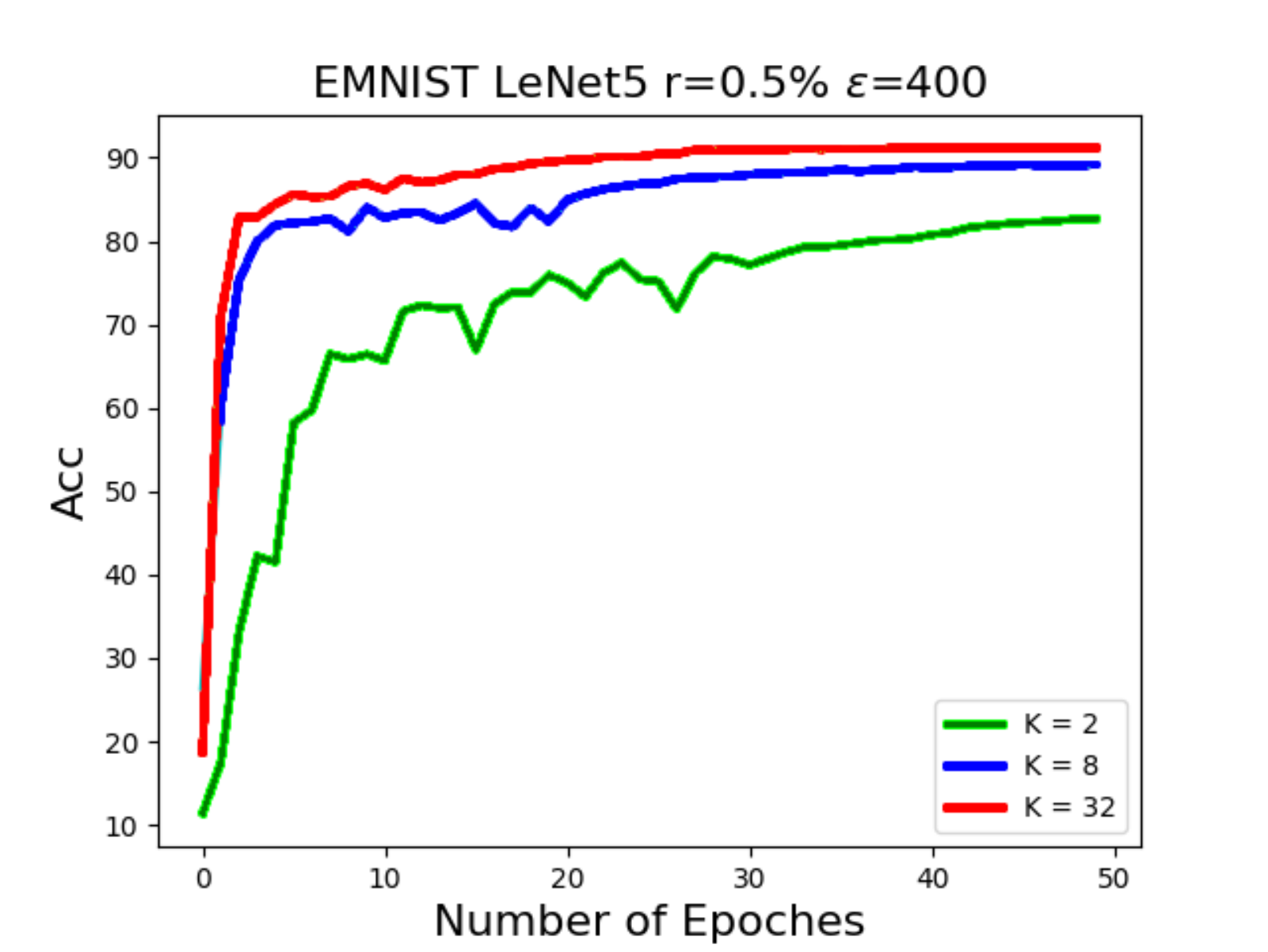}
		\includegraphics[width=45mm]{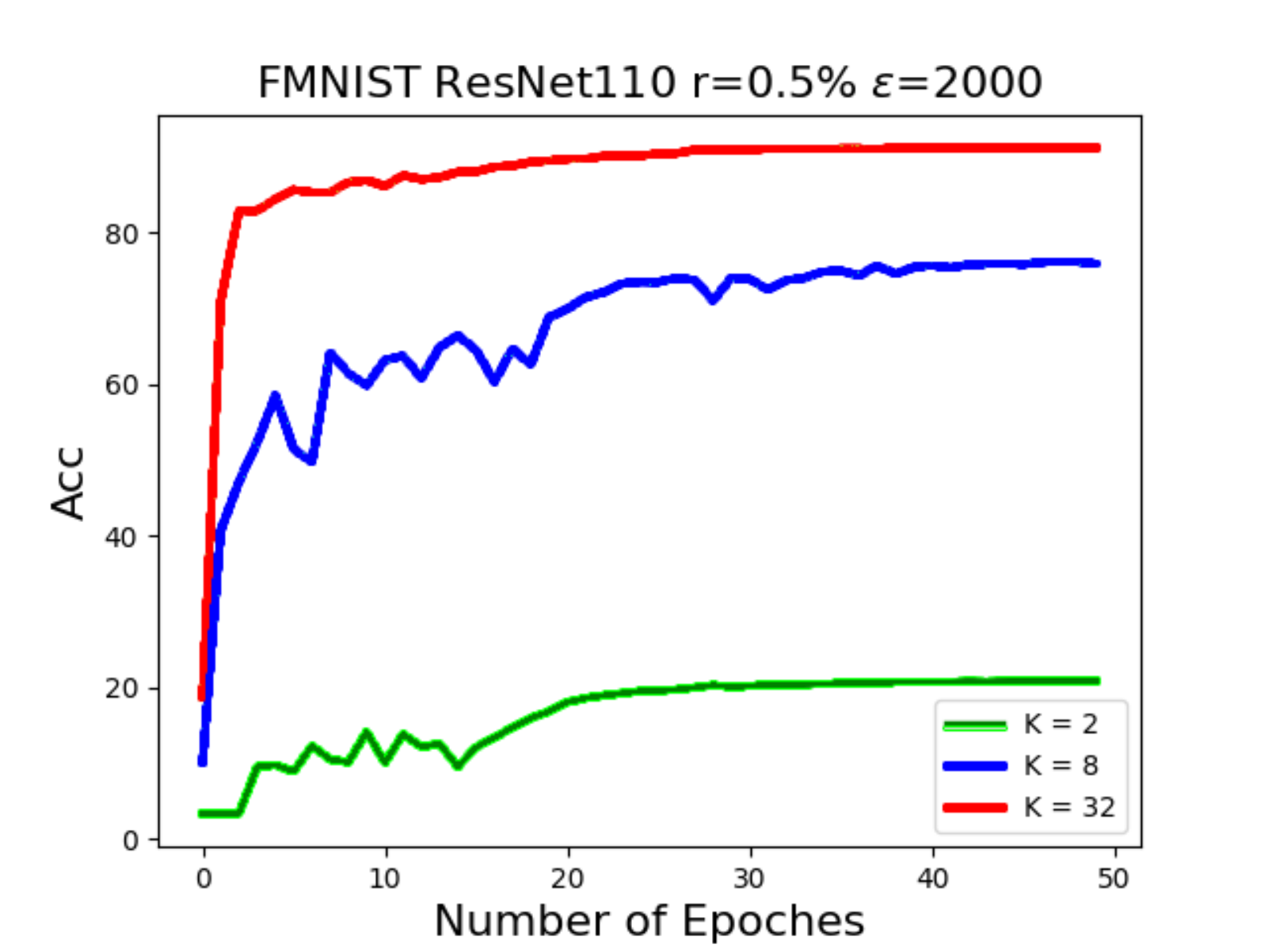}
	\end{tabular}
	\caption{Study on the impact of communication constraints}
	\label{fig: 1}
\end{figure*}
\section{Ablation Studies} \label{ablation}
\subsection{Impact of communication constraints} We evaluate the performance of trained models using the test set accuracy. The privacy budget is chosen as $ \epsilon = 400 $ for LeNet models and $ \epsilon = 2000 $ for ResNet models. Sampling ratio is fixed at $ r = 0.5\% $. We vary the quantization level in the range $ \{2, 8, 32\} $ which corresponds to using $ \{1, 3, 5\} $ bits per client per dimension. We did not use the adaptive gradient norm shrinking strategy and kept the norm bound as $ U = 10 $ throughout the training procedure across all the setups. The results are plotted in Figure \ref{fig: 1}.

\begin{figure}
	\centering
	\begin{tabular}{ccc}
		\includegraphics[width=45mm]{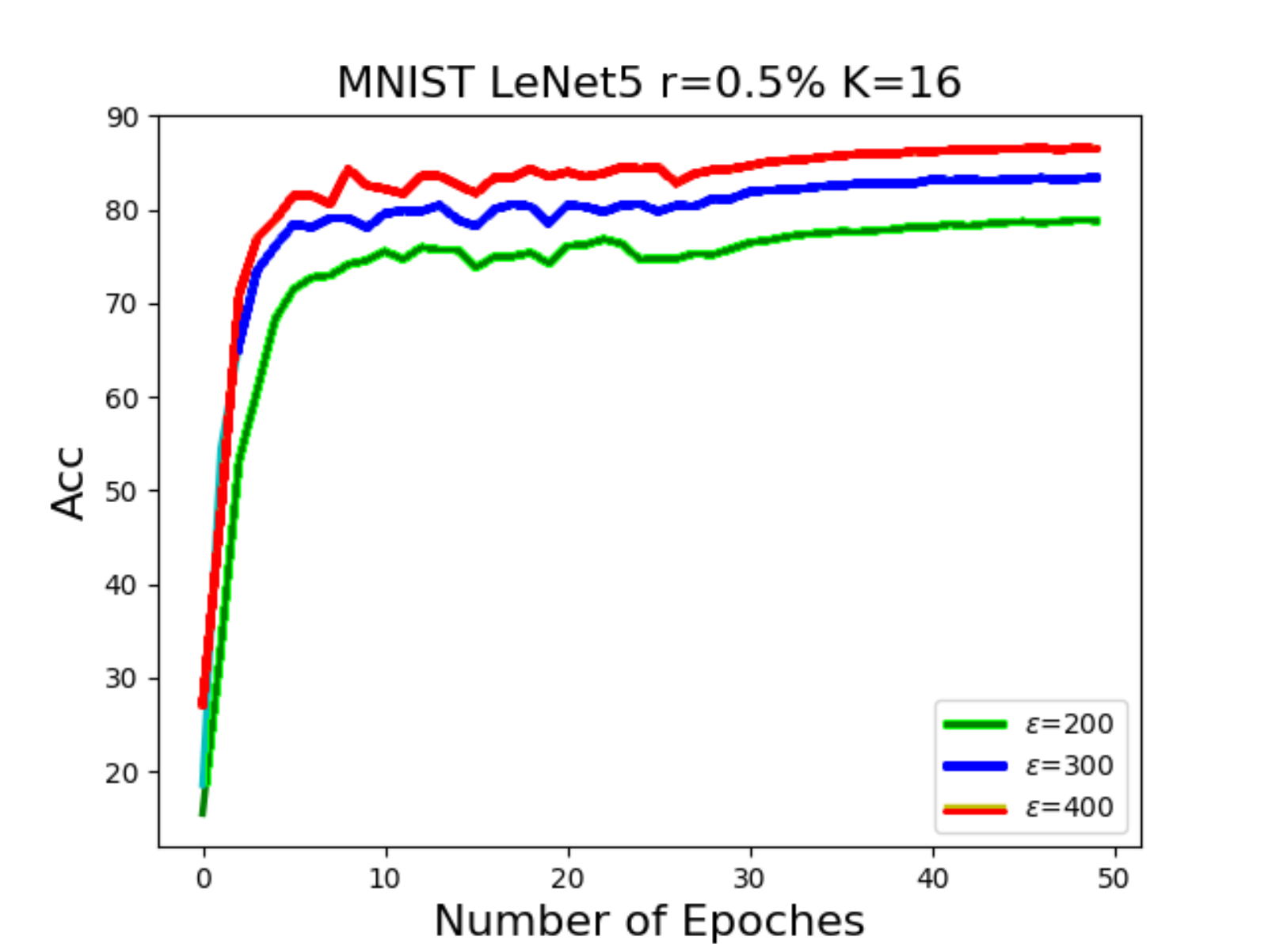} &
		\includegraphics[width=45mm]{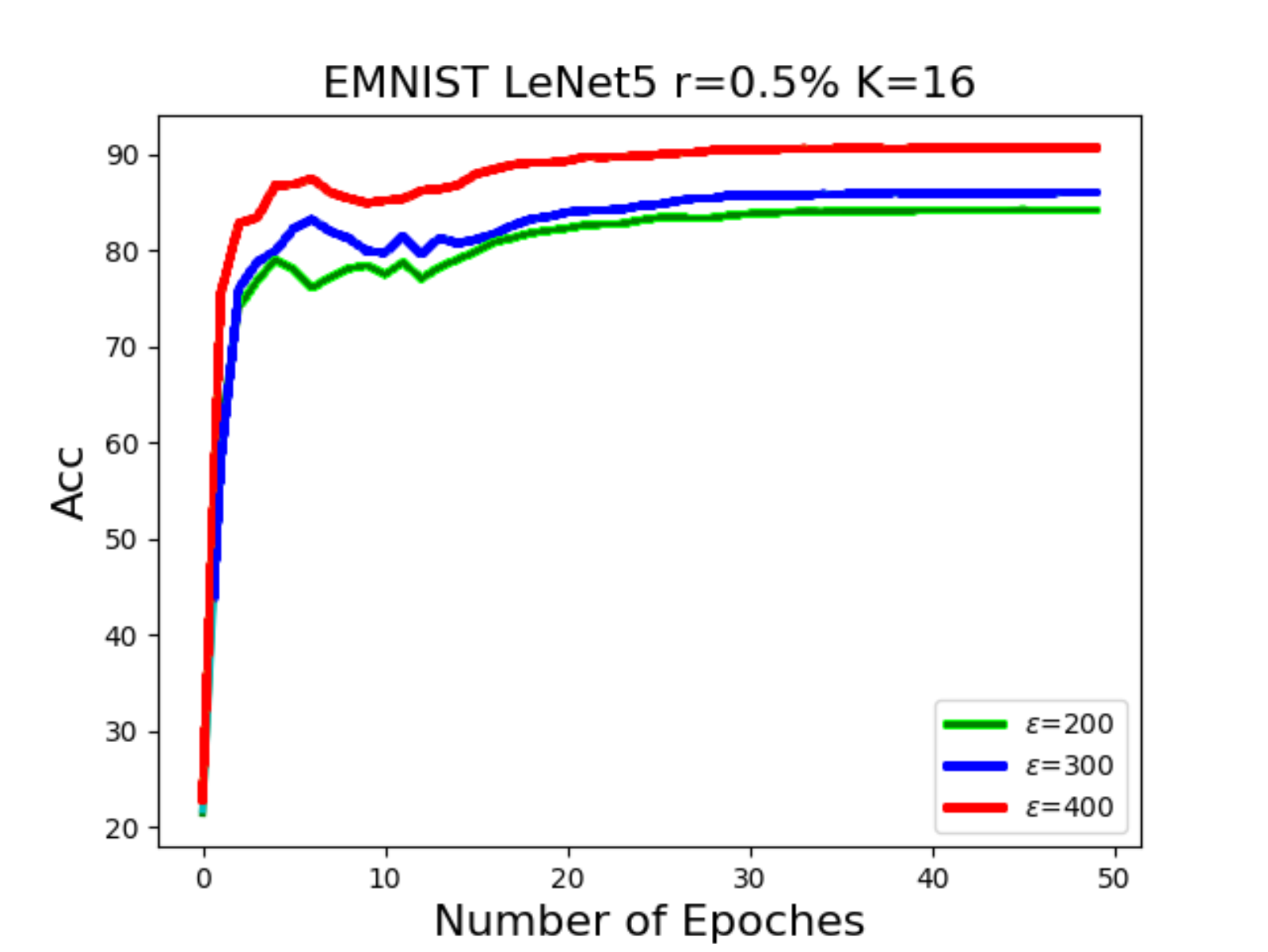} &
		\includegraphics[width=45mm]{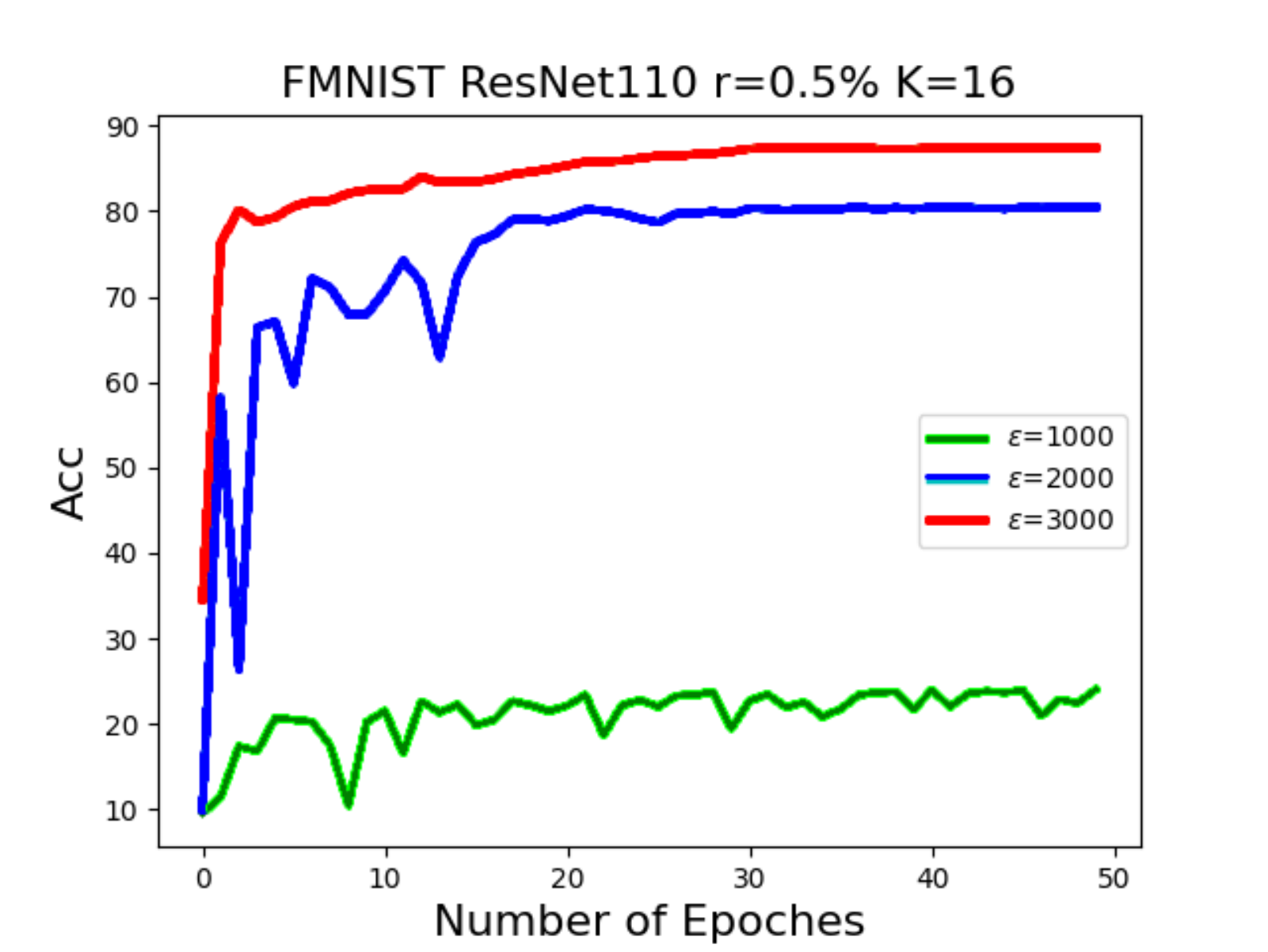}
	\end{tabular}
	\caption{Study on the impact of privacy constraints}
	\label{fig: 4}
\end{figure}
The results indicate that while $ 1 $-bit sqSGD is efficient in terms of communication cost, the quantization error results in significant degradation in model performance, both in the final accuracy after model convergence, and in the speed of convergence. A reasonable quantization level (i.e. larger than $ 3 $ bits) yields competitive performance. \par
\subsection{Impact of privacy constraints} Next we investigate the performance loss due to privacy constraints. We fix the sampling rate at  $ r = 0.5\% $ and quantization level at $ K=16 $. For LeNet-5, we select $ \epsilon $ from $ \{200, 300, 400\} $, for ResNet-110, we select $ \epsilon $ from $ \{1000, 2000, 3000\} $. Gradient norm bounds are kept fixed at $ U = 10 $. The results are plotted in Figure \ref{fig: 4}, which imply that with limited communication, we need high privacy budgets to ensure successful training on large models.
\begin{figure}
	\centering
	\includegraphics[width=50mm]{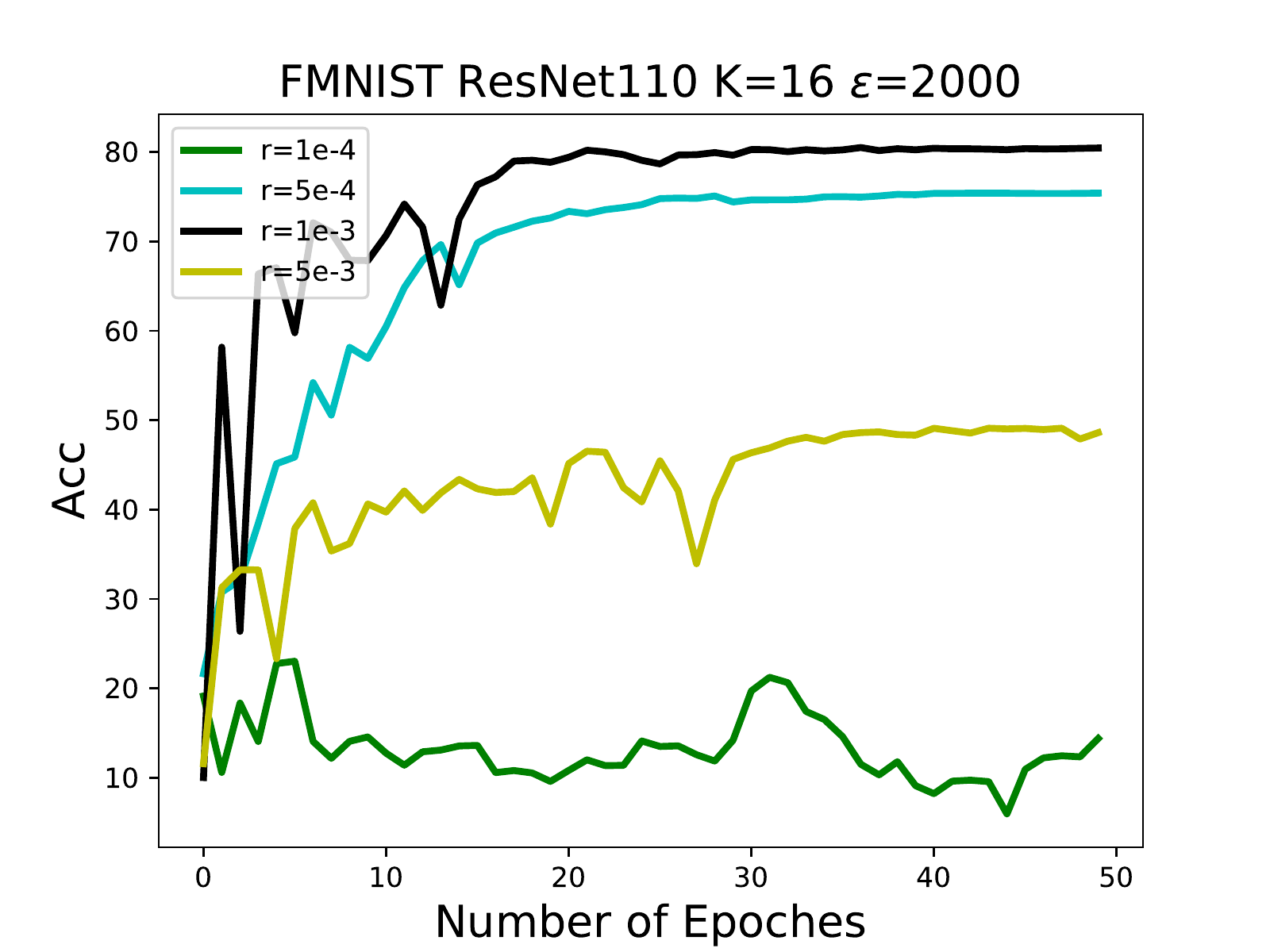}
	\caption{Study on the effect of gradient subsampling}
	\label{fig: 2}
\end{figure}

\subsection{Impact of gradient subsampling} \label{subsample}
In this experiment, we investigate the effect of subsampling under the setup of training a ResNet110 model on the FMNIST dataset. We fix the privacy level at $ \epsilon = 2000 $ and quantization level at $ K = 16 $. We vary the subsampling ratio from the set $ \{1, 5, 10, 50\} \times 10^{-3} $. Gradient norm bounds are kept fixed at $ U = 10 $. The results are plotted in Figure \ref{fig: 2}. It could be seen from the plot that using a high sampling ratio not only severely hurts the training performance, but also incurs more communication. It is thus necessary to perform subsampling. However, using a very low subsampling ratio also causes training failure. This phenomenon will be further explored in the next experiment. \par 

\begin{figure}
	\centering
	\includegraphics[width=50mm]{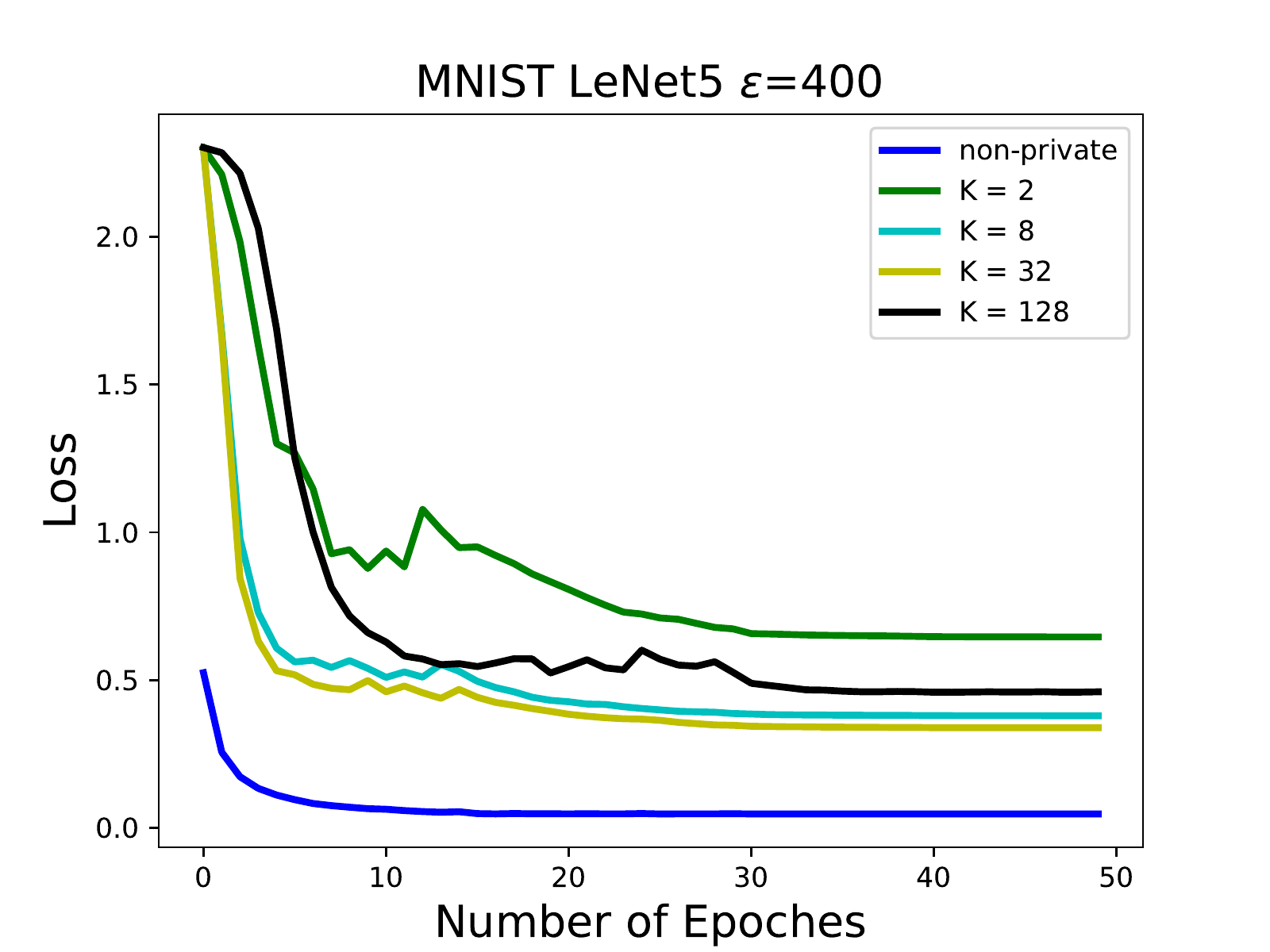}
	\caption{Study on the trade-off between quantization and sampling with fixed communication level}
	\label{fig: 3}
\end{figure}

\subsection{Trade-off between quantization and sampling}
In this experiment, we study the trade-off between quantization and sampling under the setup of training a LeNet-5 model on the MNIST dataset. We fix the privacy level at $ \epsilon = 400 $, and fix the total communication cost per client, measured using the product $ r\log_2 K $. We vary the quantization level in the range $ \{2, 8, 32, 128\} $ which corresponds to using $ \{1, 3, 5, 7\} $ bits per client per dimension, and the sampling rate is adjusted accordingly. Gradient norm bounds are kept fixed at $ U = 10 $. The results are shown in Figure \ref{fig: 3}. The results suggest that increasing the quantization level may not monotonically increase training performance. This is mainly due to the random subsample scheme of sqSGD, under which the structure of gradients is not fully explored.

%
%
%

%




\end{document}